%% file: AAAI2025.tex
\documentclass[letterpaper]{article} 
\usepackage{aaai25}  
\usepackage{times}  
\usepackage{helvet}  
\usepackage{courier}  
\usepackage[hyphens]{url}  
\usepackage{graphicx} 
\usepackage{natbib}  
\usepackage{caption} 
\usepackage{algorithm}
\usepackage{algorithmic}
\usepackage{amsfonts}
\usepackage{amsmath}
\usepackage{multirow}
\usepackage{pifont}
\newcommand{\cmark}{\ding{51}}%
\newcommand{\xmark}{\ding{55}}%
\usepackage{array}
\newcolumntype{H}{>{\setbox0=\hbox\bgroup}c<{\egroup}@{}}
\usepackage{xcolor} 
\usepackage{soul}   
\usepackage{marvosym}
\definecolor{myblue}{RGB}{0, 0, 255} 
\definecolor{myred}{RGB}{255, 0, 0}  
\definecolor{lightgray}{gray}{0.7}

\newcommand{\best}[1]{\textbf{#1}}
\newcommand{\secondbest}[1]{{\textit{#1}}}
\usepackage{newfloat}
\usepackage{listings}
\DeclareCaptionStyle{ruled}{labelfont=normalfont,labelsep=colon,strut=off} 
\floatstyle{ruled}
\newfloat{listing}{tb}{lst}{}
\floatname{listing}{Listing}
\title{Medical Manifestation-Aware De-Identification}
\author{
    Yuan Tian\textsuperscript{\rm 1},
    Shuo Wang\textsuperscript{\rm 2},
    Guangtao Zhai\textsuperscript{\rm 2}\thanks{Corresponding Author.}
}
\affiliations{
    \textsuperscript{\rm 1} Shanghai AI Laboratory, \\
    \textsuperscript{\rm 2} Institute of Image Communication and Network Engineering, Shanghai Jiao Tong University \\
    tianyuan168326@outlook.com, zhaiguangtao@sjtu.edu.cn
}

\usepackage{bibentry}

\begin{document}

\maketitle

\begin{abstract}
Face de-identification (DeID) has been widely studied for common scenes, but remains under-researched for medical scenes, mostly due to the lack of large-scale patient face datasets. In this paper, we release \textbf{MeMa}, consisting of over 40,000 photo-realistic patient faces. MeMa is re-generated from massive real patient photos. By carefully modulating the generation and data-filtering procedures, MeMa avoids breaching real patient privacy, while ensuring rich and plausible medical manifestations. We recruit expert clinicians to annotate MeMa with both coarse- and fine-grained labels, building the first medical-scene DeID benchmark. Additionally, we propose a baseline approach for this new medical-aware DeID task, by integrating data-driven medical semantic priors into the DeID procedure. Despite its conciseness and simplicity, our approach substantially outperforms previous ones.
\end{abstract}

\begin{links}
 \link{Dataset and Code}{https://github.com/tianyuan168326/MeMa-Pytorch}
\end{links}

\vspace{-2mm}
\section{Introduction}
The public sharing of large-scale image datasets has facilitated the rapid progress in Artificial Intelligence (AI).
However, this also poses great privacy concerns, especially for facial images, which are widely used for identity authentication.
To address this issue, many de-identification (DeID) algorithms~\cite{cao2021personalized,maximov2020ciagan,gu2020password,li2023riddle,cai2024disguise} have been continuously proposed for protecting the facial identity, achieving promising results on common-scene facial datasets~\cite{karras2019style,karras2017progressive}.
\begin{figure}[t]
	\centering
	\centering
	\setlength{\tabcolsep}{1pt} 
	\begin{tabular}{cc}
		
		\includegraphics[width=0.485\linewidth]{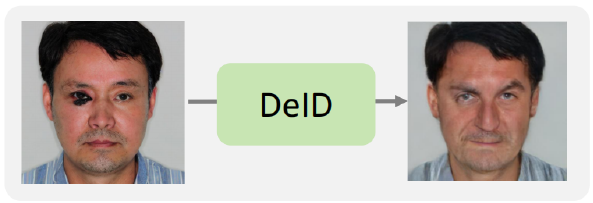}&
		\includegraphics[width=0.485\linewidth]{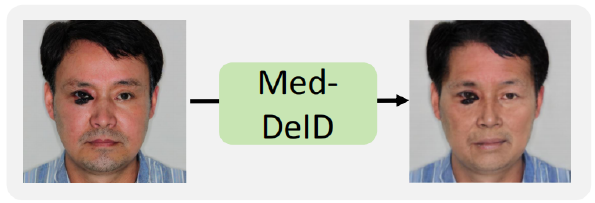}\\
		(a) DeID &
		(b) Med-DeID \\
		\includegraphics[width=0.48\linewidth]{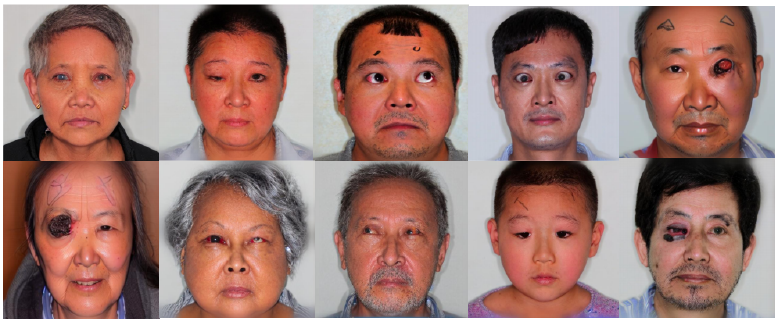} &
		\includegraphics[width=0.48\linewidth]{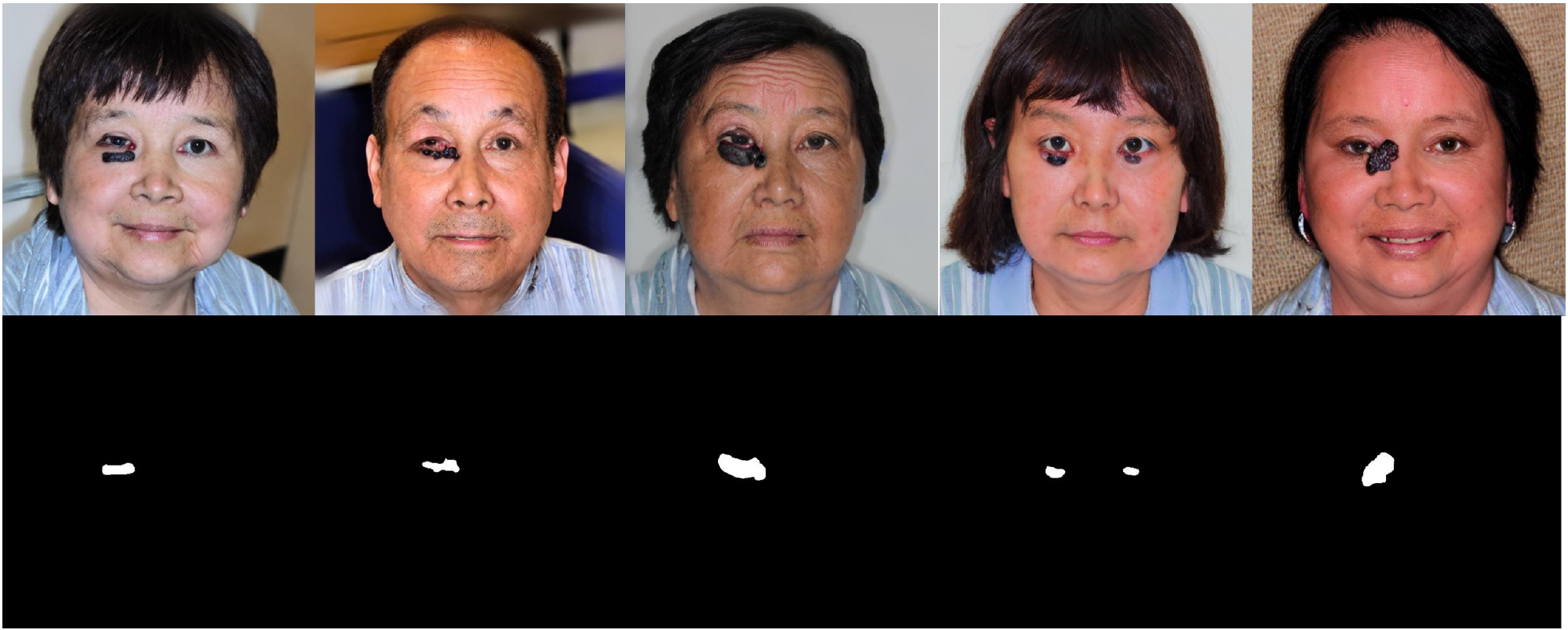}\\
		(c) MeMa &
		(d) MeMa-Seg \\
		
	\end{tabular}
		\vspace{-2mm}
	\caption{
		(a) Common DeID approaches, focus on removing identity.
		(b) Our medical-aware DeID (Med-DeID), also considers preserving the diagnosis-necessary medical information.
		(c) Our MeMa, a large-scale patient face dataset.
		(d) Our MeMa-Seg, the tumor segmentation subset of MeMa.
	}
	\vspace{-5mm}
	\label{fig:mema_dataset}
\end{figure}

However, rare researches are conducted for the medical scenes, although patient privacy leakage is a big concern in the medical AI era~\cite{price2019privacy}.
Research on medical-aware DeID (Med-DeID) mainly faces two obstacles.
\textit{First}, there are few medical-scene facial datasets available, due to the difficulty in accessing patients compared to healthy individuals. Moreover, it is often not acceptable to package real patient faces as datasets and make them publicly downloadable.
\textit{Second}, the current DeID approaches may not be appropriate for protecting medical facial images, due to not particularly preserving the disease manifestations of the origin image. This leads to the lost of diagnosis-necessary disease signs, deteriorating the medical utilities.

In this paper, we release a Medical Manifestation-rich patient face dataset, termed \textbf{MeMa}, containing over 40,000 photo-realistic virtual patient images. To construct MeMa, we obtained permission from the hospital's medical ethics committee to photograph patients. Then, these patient photos are annotated by expert physicians, before being used to train a specialized generative model. By carefully modulating the sampling procedure of the generative model and filtering the generated data, we created a diverse, high-quality, and real-world-like patient face dataset.
 \begin{figure*}[!htbp]
	\centering
	\begin{tabular}{cccc}
		\setlength{\tabcolsep}{0pt} 
		\includegraphics[width=0.29\linewidth]{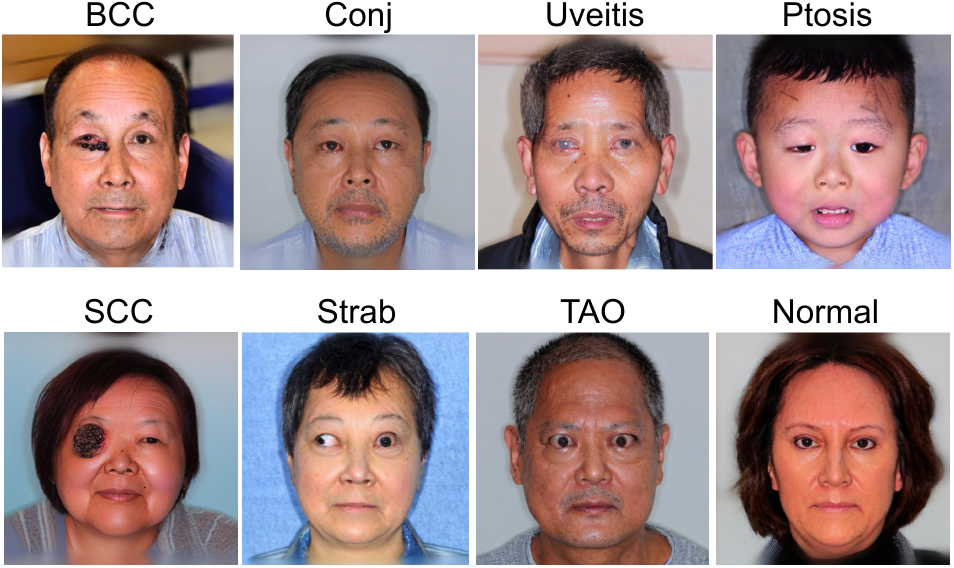}&
		\includegraphics[width=0.20\linewidth]{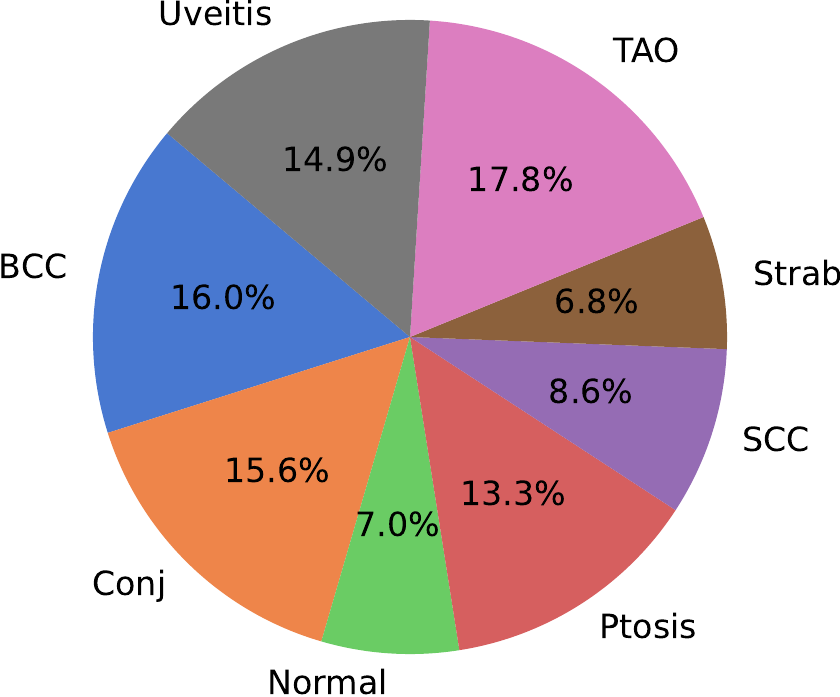}&
		\includegraphics[width=0.20\linewidth]{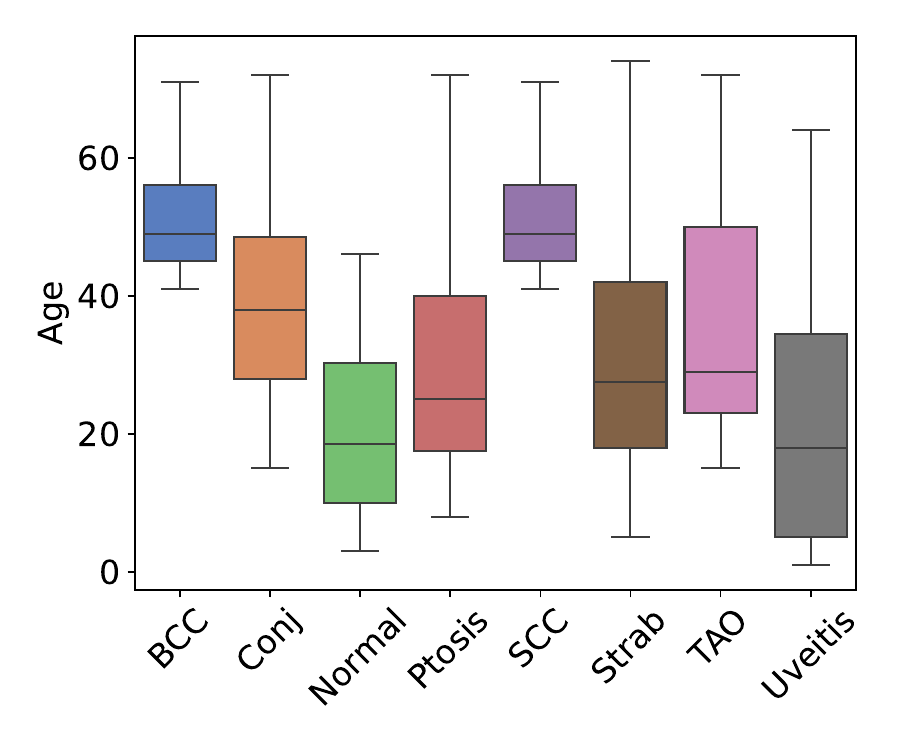} &
		\includegraphics[width=0.20\linewidth]{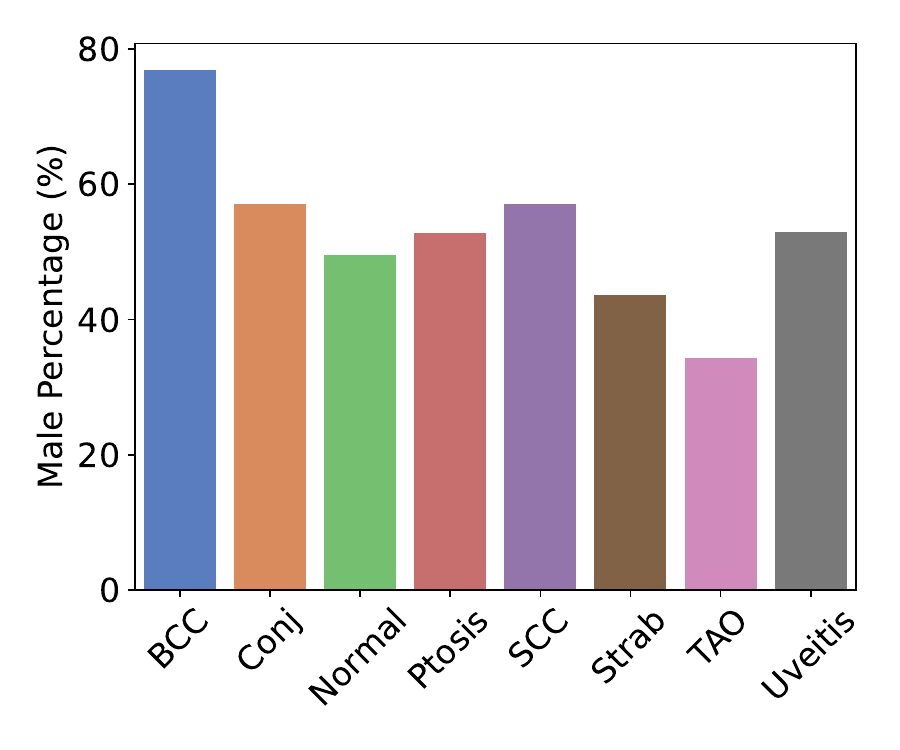}\\
		(a) Disease Example&
		(b) Patient distribution&
		(c) Age distribution&
		(d) Gender distribution
	\end{tabular}
	\vspace{-2mm}
	\caption{Examples and the distribution characteristics of the proposed MeMa dataset.} 
	\vspace{-4mm}
	\label{fig:dataset_distribution}
\end{figure*}

Furthermore, we propose a baseline medical semantics-preserved DeID approach, termed MedSem-DeID, to eliminate the patient identity from the facial image, at the premise of preserving the medical utility.
Concretely, we first condense the rich medical priors within the MeMa into a medical semantics encoder, and then adopt it to (1) enhance the medical knowledge of the features within the DeID pipeline, and (2) minimize the medical-aware distortion of the de-identified images.
Despite its simplicity, our approach easily outperforms previous DeID approaches for medical scenes, thanks to the rich medical manifestation knowledge embodied in the MeMa dataset.
Our main contributions are:
\begin{itemize}
	\item We release, to the best of our knowledge, the first large-scale patient face dataset of rich medical manifestations, \textbf{MeMa}, which is expected to facilitate research in the field of medical-scene privacy protection.
	\item We propose a baseline approach for this novel medical DeID problem, which particularly preserves the disease signs during the DeID procedure, by making full use of the rich medical priors within MeMa.
	\item We build the first medical-scene DeID benchmark, by comprehensively evaluating the proposed baseline and other recent DeID approaches on MeMa. Our approach is consistently superior in various aspects.
\end{itemize}

\section{Related Work}
\vspace{-1mm}
\noindent \textbf{Facial Datasets.}
Amounts of large-scale face datasets~\cite{karras2019style,karras2017progressive,huang2008labeled,karkkainen2021fairface} have been proposed, but they primarily feature healthy individuals, limiting their use for medical-scene DeID.
In contrast, we introduce a large-scale patient face dataset with rich medical manifestations. Our dataset includes annotations for disease type and lesion masks, facilitating face DeID field in medical scenes.

\noindent \textbf{Face De-identification.}
Early De-ID methods~\cite{gross2006model,jourabloo2015attribute} used the K-same algorithm. Recent approaches~\cite{hukkelaas2019deepprivacy,maximov2020ciagan} leverage generative models to remove facial identity, while often compromising utility. More recent methods~\cite{wen2023divide,cai2024disguise,ren2018learning} aim to preserve more facial attributes and better serve common utilities such as gaze detection~\cite{hu2020dgaze} and image/video recognition~\cite{kong2022human,tian2022ean,tian2020self,tian2021self,tian2019video,yan2023dhbe,gao2024imofc,tan2024low}, but not specifically medical signs~\cite{mohsenin2012ocular,chen2024cross}. In contrast, our approach leverages medical manifestation representations learned from real patient photos, preserving medical attributes during DeID. Additionally, it is reversible, similar to~\cite{gu2020password,cao2021personalized,li2023riddle}, enabling reversal for medical audits.

\noindent \textbf{Semantic Representation.}
Effectively modeling semantic information is crucial for modifying facial images, while maintaining perceptual quality~\cite{min2024perceptual,yi2021attention,duan2022develop,chen2024gaia,li2024r,gao2022two,gao2021towards,yi2024no} and preserving medical utility. Previous approaches have leveraged contrastive learning~\cite{he2020momentum,tian2024coding,tian2023clsa} and masked image modeling~\cite{he2022masked,tian2023non,tian2024smc++} for self-supervised learning of image semantics. Recent studies have shown that pre-trained visual foundation models, such as stable diffusion~\cite{rombach2022high}, exhibit even stronger semantic representations~\cite{zhang2024vision,tian2025free,hedlin2024unsupervised,li2024sd4match}. 
In this work, we present the first adaptation of diffusion model-extracted semantics to the medical DeID problem.

\noindent \textbf{Medical-scene Face Privacy Protection.}
Progress on this problem has been slow, often relying on simple methods like blurring or replacing faces with 3D masks~\cite{yang2022digital}, which discard critical disease signs. The progress gap is attributed to the lack of large-scale medical-scene facial datasets. Our work aims to address this gap.

\section{Approach}
We first build a new patient face dataset, termed \textbf{MeMa}. It addresses the lack of medical-scene facial datasets. MeMa is synthesized from real patient photos. Its synthetic nature avoids potential ethical problems. Expert physicians recognize its validity. Further, we propose a baseline model for the medical-aware facial DeID (Med-DeID) problem.
\begin{figure}[!thbp]
	\vspace{-0mm}
	\centering
		\begin{tabular}{c}
			\includegraphics[width=0.99\columnwidth]{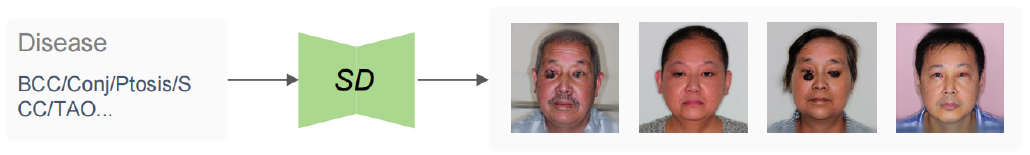} \\
			(a)\vspace{-0mm}  \\
			\includegraphics[width=0.99\columnwidth]{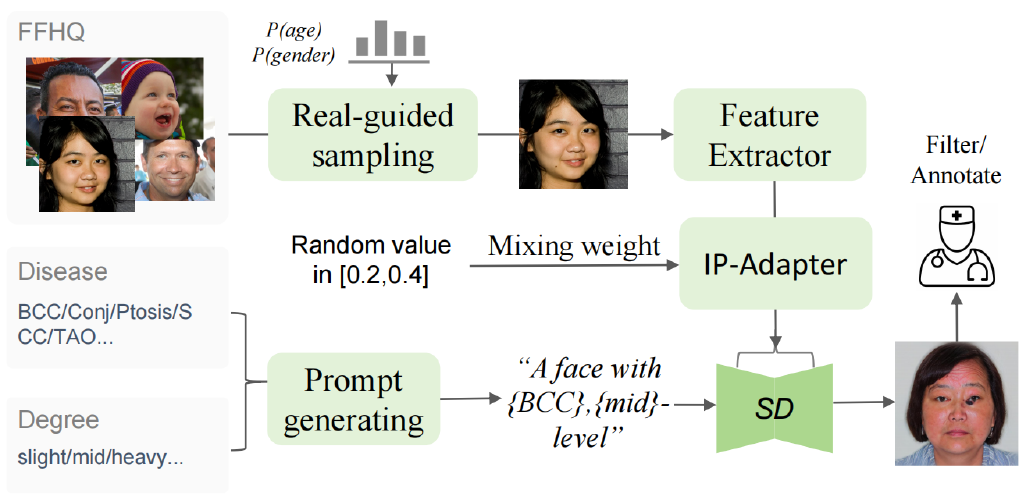}\vspace{-4mm} \\
			(b) 
	\end{tabular}
	\vspace{-2mm}
	\caption{MeMa building pipeline. (a) Training patient face generation model on real patient data. (b) Rich-condition patient face sampling.
		$P({age})$ and $P({gender})$ denote the age and gender distributions, which are statistically derived from the real patients.
		`SD' denotes the stable diffusion model.
	} 
	\vspace{-6mm}
	\label{fig:dataset_build}
\end{figure}

	\vspace{-1mm}
\subsection{MeMa Dataset} 
\label{sec:method_mema}
The overview of MeMa is shown in Fig.~\ref{fig:dataset_distribution}, which consists of 42,307 synthetic patient face images. MeMa closely mimics real patients in both visual appearance and statistics. Patient age and gender are estimated using the DeepFace framework~\cite{serengil2020deepface,serengil2021lightface}.
We describe the main steps for assembling MeMa as follows.

\noindent \textbf{Disease Categories:}
We take the eye clinic as an exemplar scene, since most eye diseases show typical external facial manifestations. In our study, we included patients with seven eye diseases. These are Basal Cell Carcinoma (BCC), Conjunctivitis (Conj), Uveitis, Ptosis, Squamous Cell Carcinoma (SCC), Strabismus (Strab), and Thyroid Associated Ophthalmopathy (TAO). We also included clinically Normal cases. Examples are shown in Fig.~\ref{fig:dataset_distribution}(a).
The detailed manifestations of the above diseases can be found in the MSD medical manual~\cite{msdmanuals2024}.

\noindent \textbf{Real Patient Data Collection:}
We collected 39,323 photos of 12,467 real patients. They attended the Eye Clinic at Shanghai Ninth People's Hospital(SNPH) between January 2020 and June 2023. The photo-taking procedure was approved by the hospital's ethics committee. The patients' diagnosis results were collected from their medical records.

\noindent \textbf{Generating MeMa from Real Data:}
As shown in Fig.~\ref{fig:dataset_build}, we first train a medical-aware generative model with the collected patient data.
Then, we sample the virtual patients from the model by using proper conditions, aiming to generate \textit{safe} and \textit{diverse} samples.
Finally, we recruit expert physicians to filter the images of bad medical quality, then annotate the filtered images.
The steps are detailed as follows.

\begin{figure}[!tbp]
	\centering
	\includegraphics[width=0.95\columnwidth]{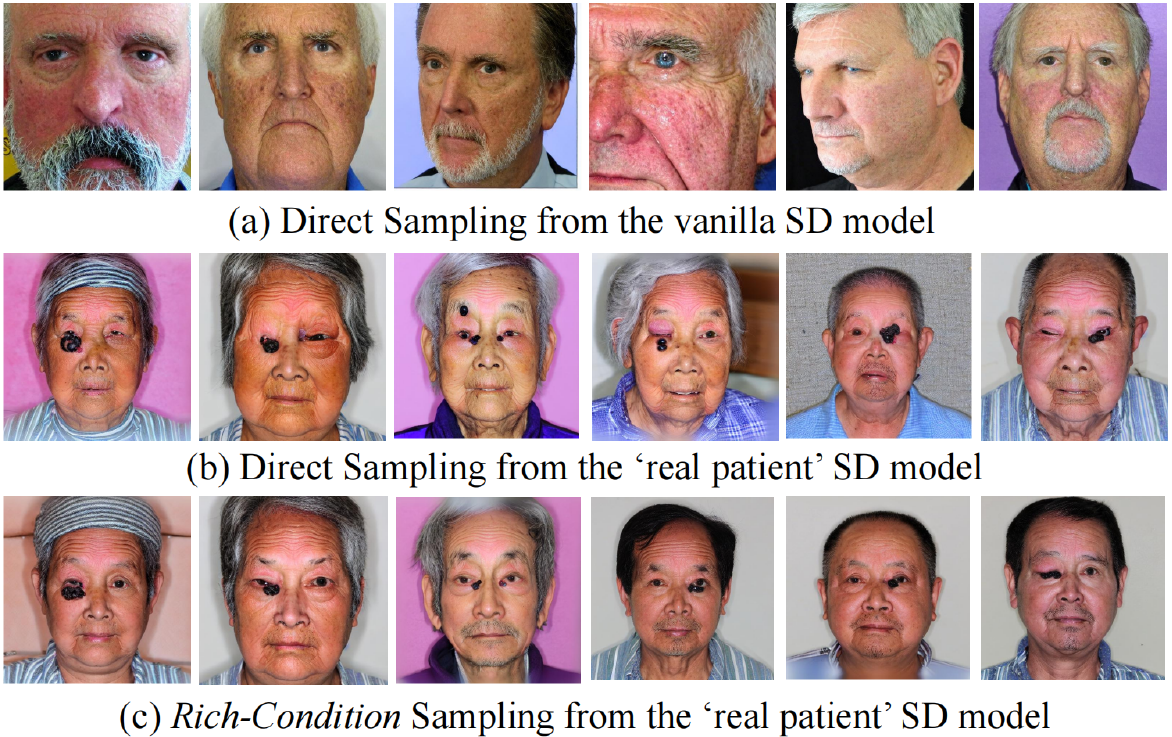} 
	\vspace{-2mm}
	\caption{Comparison of different image generation strategies. We take the Basal Cell Carcinoma (BCC) disease as an example.
		`SD' denotes the Stable Diffusion.
	}
	\vspace{-3mm}
	\label{fig:gen_variant_model}
\end{figure}

\textit{Step1: Medical-aware Generative Model Training:} 
We first translate the disease type into the prompt caption `A face, eye with \{disease name\}'.
With the paired data of the real patient photographs and the disease type caption, we fine-tune the diffusion model~\cite{rombach2022high}, producing the patient face generation model.
As compared in Fig.~\ref{fig:gen_variant_model} (a) and (b), after fine-tuning the SD model on our real patient dataset, the generated image shows typical medical signs and manifestations, while the vanilla SD model can not effectively generate images with reasonable medical manifestations, due to its limited medical knowledge.

\textit{Step2: Rich-Condition Patient Face Synthesis:}
Directly sampling from the real-patient generation model with the simple prompt `A face, eye with \{disease name\}' is not enough, which shows two problems. First, {identity leakage}: the identity of most sampled patients can be found in the training dataset, potentially leaking the privacy of real patients. Second, {mode collapse}: the samples tend to be less diverse, with collapsed medical manifestation modes.

To address the \textit{identity leakage} problem, we propose injecting facial attributes from public faces into the generation process. Specifically, we randomly sample face images from the FFHQ dataset and use the IP-Adapter~\cite{ye2023ip} to inject these attributes. As shown in Tab.~\ref{tab:dataset_reduce_privacy}, this substantially reduces the average identity leakage percentage from 71.8\% to 1.27\%, when being evaluated with multiple face recognition models, \textit{i.e.}, SphereFace~\cite{liu2017sphereface}, ArcFace~\cite{deng2019arcface}, and CosFace~\cite{wang2018cosface}.

\begin{table}[!thbp]
	\setlength{\tabcolsep}{3pt} 
	\centering
	\begin{tabular}{c|c|c|c|c}
		\hline
		 & SphereFace& ArcFace&CosFace&Average \\
		\hline
		Direct Sample& 73.45\% & 76.72\% & 65.34\% &71.83\% \\
		\hline
		Face Injection& 1.62\% 	& 0.97\% & 1.24\% & 1.27\%  \\
		\hline
	\end{tabular}
		\vspace{-2mm}
	\caption{
		Effectiveness of injecting public face attribute for reducing the identity leakage percentage.
	} 
	\vspace{-3mm}
	\label{tab:dataset_reduce_privacy}
\end{table}

To mitigate the \textit{mode collapse} problem, we first randomly sample the public face injection weight from the range $[0.2, 0.4]$, instead of using a fixed weight. This leads to better feature fusion flexibility and improves output diversity.
Second, we enhance the text prompt with severity descriptions, e.g., `A face, eye with \{disease name\}, \{slight/mid/heavy\}-level'. This further improves diversity, even though no disease severity is annotated in the collected patient captions. The reason may be that the base SD model has learned a large dictionary of word semantics and can automatically connect the common `severity' description words to the image generation process.
As compared in Fig.~\ref{fig:gen_variant_model} (b) and (c), with rich conditions injected, both the quality and diversity of the generated images are substantially improved.

To ensure the generated dataset's statistical characteristics match those of real patients, we calculate the distributions of real patient disease types, ages, and genders. We control the generated images to follow the above distributions.
To control the disease type, we simply modify the disease name of the prompt.
To control the age and gender of the generated images, we label FFHQ images using an age and gender estimation model~\cite{serengil2020deepface}, then select images based on this metadata for attribute injection. As shown in Tab.~\ref{tab:dataset_reduce_dist_gap}, the real distribution-guided sampling strategy produces a dataset with similar statistical characteristics to real patients.

\begin{table}[!tbp]
	\setlength{\tabcolsep}{8
		pt} 
	\centering
	\begin{tabular}{c|c|c|c}
		\hline
		Sampling Strategy & Disease & Age &Gender \\
		\hline
		Random sampling & 0.256 &0.143 & 0.157 \\
		\hline
		Real-guided sampling& \textbf{0.003} & \textbf{0.002} & \textbf{0.002} \\
		\hline
	\end{tabular}
	\vspace{-2mm}
	\caption{
		Wasserstein distance~\cite{ruschendorf1985wasserstein} between the generated and the real patient image distributions. Smaller distance indicates more real-world alike generation.
	} 
	\vspace{-6mm}
	\label{tab:dataset_reduce_dist_gap}
\end{table}

\textit{Step3: Filtering and Annotation:}
After generating the images, we remove those with small identity feature distance to the original real patient set, ensuring the privacy of the real patients will not be leaked. Then, the physicians filter out the images with low medical utility quality. Finally, these physicians label the per-image disease information of the filtered dataset.
Moreover, considering that lesion segmentation is another representative medical imaging task.
We also ask the physicians to segment the tumor mask of the subset SCC images, producing the MeMa-Seg subset. The annotation procedure is assisted by the SAM model~\cite{kirillov2023segment} and then refined by the physicians.

	\vspace{-1mm}
\subsection{A Baseline Approach for Med-DeID}
We propose a baseline approach to incorporate the rich medical manifestation knowledge within MeMa into the DeID procedure, which consists of two sub-modules: medical semantics encoding and medical semantics-preserved DeID.

\noindent \textbf{Medical Semantics Encoding.}
The Med-DeID task requires preserving as much medical information as possible while obfuscating other identifying details. This necessitates a semantic encoder that recognizes local medical semantics.

Motivated by the strength of diffusion models in extracting fine-grained local semantics~\cite{tian2024diffuse,tang2023emergent}, we train another diffusion model on the proposed MeMa dataset to learn the medical semantics.
We adopt its first several blocks as the medical encoder $Enc_{{med}}$, instead of the whole network, for reducing the computational cost.

It should be mentioned that the roles of the diffusion models in the previous section and here are fundamentally different: the previous one is for high-quality image generation, whereas the one here is for extracting rich medical semantics.
Our approach is very flexible, and the semantic encoder can be other choices, as analyzed in the experiment section.

\begin{figure}[!htbp]
	\centering
	\includegraphics[width=0.99\columnwidth]{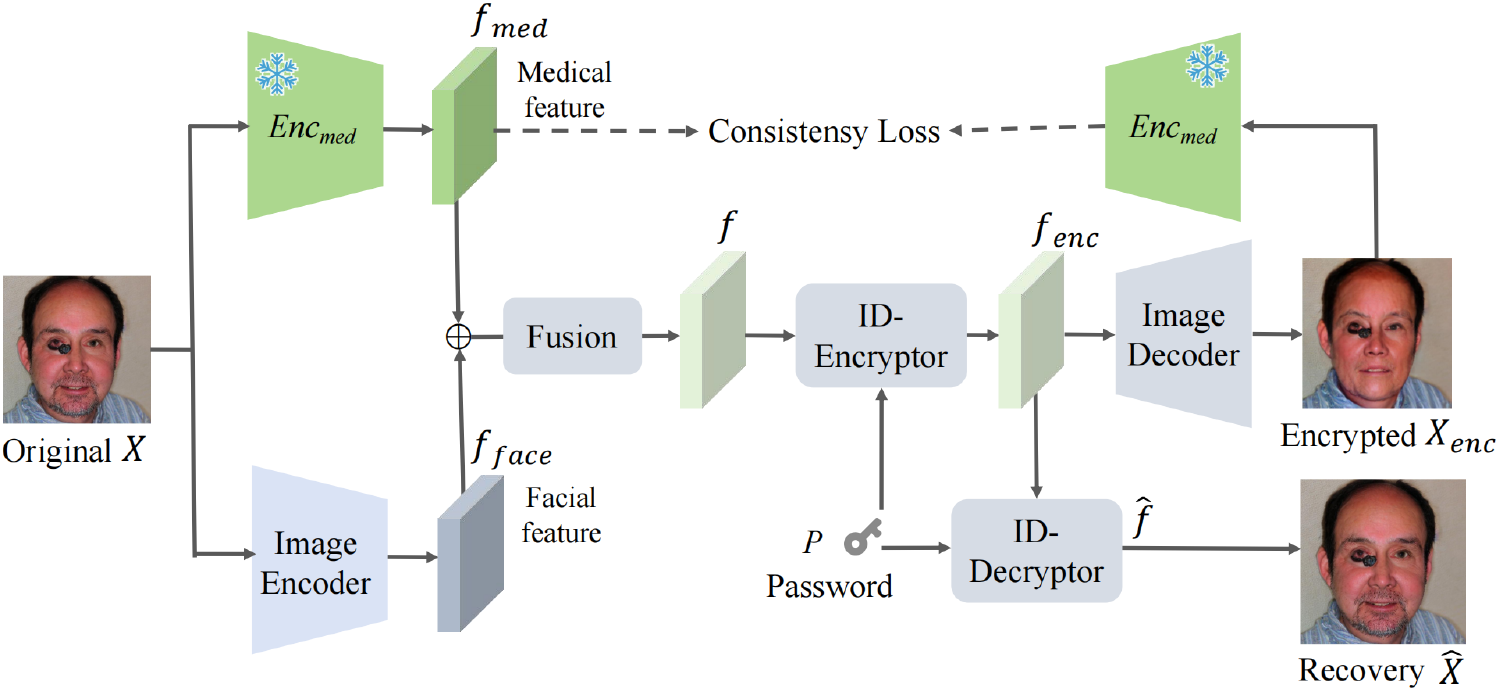}
	\vspace{-2mm}
	\caption{Overview of the proposed baseline model MedSem-DeID.
		The snow icon indicates the $Enc_{\text{med}}$ is frozen during training DeID networks.
		The image decoder after the ID-decryptor is omitted for briefness.
		$\oplus$ denotes the channel-wise concatenation operation.
		 }
		 \vspace{-3mm}
	\label{fig:method}
\end{figure}

\noindent \textbf{Medical Semantics-Preserved DeID (MedSem-DeID).}
As illustrated in Fig.~\ref{fig:method}, our approach leverages the medical encoder $Enc_{{sem}}$ to inject medical knowledge into the feature extraction procedure, as well as regularize the medical utility of the de-identified image.

Given the original image $X$, where $H$ and $W$ denote its height and width, an image encoder transforms $X$ into the facial feature $f_{{face}}\in\mathbb{R}^{512 \times \frac{H}{32} \times \frac{W}{32}}$. Meanwhile, we use $Enc_{{med}}$ to extract the medical feature $f_{{med}} \in \mathbb{R}^{320 \times \frac{H}{16} \times \frac{W}{16}}$.
The $f_{{med}}$ is downscaled and concatenated with $f_{{face}}$, passing through three consecutive residual blocks~\cite{he2016deep}, producing $f$.
Then, we employ a group of Transformer blocks~\cite{vaswani2017attention}, termed ID-Encryptor, to encrypt the ID information within $f$. Specifically, we flatten the spatial dimension of $f$, concatenate it with the password vector $P \in \mathbb{R}^{512}$, and feed the concatenated vector into ID-Encryptor, producing the encrypted feature $f_{{enc}}$.
$f_{{enc}}$ is passed through an image decoder network to result in the encrypted image $X_{{enc}}$.
Please refer to the {supplementary material} for the network architecture details.
 
In medical contexts, it is often necessary to rigorously recheck results with expert physicians on the original image. Moreover, the Good Clinical Practice (GCP) guideline~\cite{guideline2001guideline} mandates that all medical materials involved in the diagnosis process must be traceable. Therefore, we design our method to be reversible, enabling the recovery of the original image from the encrypted features. Given the original password $P$, $f_{{enc}}$ can be decrypted back to $\hat{f}$, by another group of Transformer blocks termed ID-Decryptor.
Then, $\hat{f}$ is reconstructed as the original image $\hat{X}$ by the image decoder. When an incorrect password is used, $f_{{enc}}$ is reconstructed into a wrong image $X_{{wrong}}$.

\noindent \textbf{Learning Objectives.} The learning objective of the proposed MedSem-DeID is formulated as follows, $\mathcal{L} = \mathcal{L}_{{deid}} + \mathcal{L}_{{rev-id}}+ \mathcal{L}_{{wrong}}+ \lambda_{med}\mathcal{L}_{{med}} + \lambda_{rev}\mathcal{L}_{{rev}}   + \mathcal{L}_{{GAN}}$.
$\mathcal{L}_{{deid}} = \cos(\phi(X), \phi(X_{\text{enc}})) $ enforces the identity of the encrypted image apart from the original image, where $\phi$ denotes the pre-trained identity recognition network ArcFace~\cite{deng2019arcface}, $\cos$ denotes the cosine similarity.
$\mathcal{L}_{{rev-id}} = -\cos(\phi(X), \phi(\hat{X}))$ enforces the identity of reversibly decrypted image is the same as the original image.
$\mathcal{L}_{{wrong}} = \cos(\phi(X), \phi(X_{{enc}}^{{wrong}}))$ enforces the identity of the image decrypted by the wrong password far away from the original image.
$\mathcal{L}_{{med}} = \ell_2(f_{med}, Enc_{{med}}(X_{\text{enc}}))$ facilitate the encrypted image is similar to the original image in terms of medical semantics.
$\mathcal{L}_{{rev}} = \ell_1(X, \hat{X})$ regularizes the appearance of the recovered image by right password is similar to the original one.
$\ell_1$ and $\ell_2$ denote the mean absolute error (MSE) and the mean squared error (MSE) functions, respectively.
The $\mathcal{L}_{\text{GAN}}$ is the adversarial generative network (GAN) loss, enforcing the photo-realism of all images.
$\lambda_{med}$ and $\lambda_{rec}$ denote the balancing weights.

\input{exp_medical}

\vspace{-2mm}
\section{Experiments}
\noindent \textbf{Datasets.}
{\textit{MeMa:}} the proposed {MeMa} dataset consists of 42,307 images in total, which is split into a training set (34,000 images), a hyper-parameter selection set (3,729 images), and a validation set (4,578 images). All images are labeled with the disease category.
{\textit{MeMa-Seg:}} for the BCC (basal cell carcinoma) disease type, we randomly select 600 images from the training set and 150 images from the validation set of MeMa, annotating the tumor masks for these images. This results in the {MeMa-Seg} dataset, which can be used to evaluate the fine-grained medical performance of different DeID approaches.
{\textit{Real-ECXHCSU:}}
we also collaborate with Eye Center of Xiangya Hospital of Central South University (ECXHCSU), enrolling 129 patients to conduct a real-world clinical trial. This aims to validate whether our algorithm, trained on the synthetic MeMa dataset, remains effective for real-world patients.
Moreover, ECXHCSU is geographically distant from SNPH used to develop the MeMa dataset.
This aims to further emphasize the generalization capability of our approach.

\noindent \textbf{Implementation Details.}
For training the \textit{patient face generator model}, we fine-tune Stable Diffusion v1-5~\cite{rombach2022high} using the low-rank adaptation (LoRA)~\cite{hu2021lora} technique, with the real patient data. The rank number is set to 64. We use the Adam optimizer~\cite{kingma2014adam} with \(\beta_1 = 0.9\) and \(\beta_2 = 0.99\). The learning rate starts at \(1 \times 10^{-4}\) and follows a cosine decay schedule.
The batch size is 32, and the model is trained for ten epochs. It takes about five days to train the model on a machine equipped with two Nvidia A6000 GPUs.
For training the \textit{medical semantics encoder}, we use the same training strategy as above, except that the training data comes only from the MeMa training set.
For training the \textit{MedSem-DeID model}, we use the Adam optimizer with \(\beta_1 = 0.5\) and \(\beta_2 = 0.99\). The initial learning rate is \(2 \times 10^{-4}\) and is halved after 150,000 iterations. The total iteration number is 300,000. The batch size is 16. Training takes approximately two days on a machine equipped with four Nvidia 4090 GPUs.

\noindent \textbf{Benchmark Methods.}
For DeepPrivacy~\cite{hukkelaas2019deepprivacy}, Password~\cite{gu2020password}, CIA-GAN~\cite{maximov2020ciagan},
and RiDDLE~\cite{li2023riddle}, we adopt their officially released codes and models.
For Disguise~\cite{cai2024disguise} and Personal~\cite{cao2021personalized}, we request the materials from the authors.

\noindent \textbf{Evaluation Protocol and Metrics.}
{\textit{Medical utility:}}
for the disease classification task, we fine-tune the DiNov2 model~\cite{oquab2023dinov2} on the MeMa training set. We evaluate its Top1 accuracy on the MeMa validation set processed by various DeID approaches.
For the tumor segmentation task, we use the nnU-Net~\cite{isensee2021nnu} to evaluate different methods on MeMa-Seg, adopting the Dice score~\cite{kamnitsas2017efficient} and Jaccard index~\cite{fletcher2018comparing} as metrics.
{\textit{Real-word clinical utility:}}
we recruit three physicians to manually diagnose the images in Real-ECXHCSU, that are de-identified by various DeID approaches.
Each image is diagnosed by all three physicians, and the final diagnosis is determined by a majority voting strategy.
We use Cohen's Kappa ($k$)~\cite{banerjee1999beyond} to measure the diagnosis consistency between the original and the de-identified images. $k$ is a common metric for evaluating clinical trial outcomes in the medical field.
{\textit{Identity protection:}}
following recent works~\cite{cao2021personalized,wen2023divide}, we use Euclidean distance between the identity features of de-identified and original faces, denoted as `ID-Dis', to quantitatively evaluate the effectiveness of identity protection.
Identity features are extracted by FaceNet~\cite{schroff2015facenet} trained on CASIA~\cite{yi2014learning}, FaceNet trained on VGGFace2~\cite{cao2018vggface2}, and SphereFace~\cite{liu2017sphereface}, which are not used in the training procedure. 
{\textit{Other utilities:}}
following previous methods~\cite{li2023riddle,cai2024disguise}, we adopt the Dlib~\cite{king2009dlib} and L2CS-Net~\cite{abdelrahman2023l2cs} to evaluate the landmark detection and gaze estimation performances.
{\textit{Reversibility:}}
we compare our method against the previous reversible methods, in terms of ID similarity, medical results, and visual quality of the reconstructed original image.

\input{exp_deid_commonutil}
\vspace{-3mm}
\subsection{Results}

\noindent \textbf{Medical Utility.}
As shown in Tab.~\ref{tab:result_medical_utility},
Our method achieves the best overall classification accuracy, outperforming the second-best method, Disguise, by more than 10\%.
For SCC disease, our approach outperforms DeepPrivacy, CIAGAN, Disguise, Password, Personal, and RiDDLE by 86.10\%, 73.53\%, 27.76\%, 34.67\%, 82.86\%, and 87.05\%, respectively.
On the more challenging tumor segmentation task, our approach also performs best, achieving the highest Dice (0.6775) and Jaccard (0.5453) scores.

Moreover, we train Password and Disguise models on our MeMa dataset, improving their classification accuracy to 54.67\% and 77.12\%, respectively, but still lagging behind our 86.70\%. This indicates that MeMa can enhance the efficacy of various DeID methods in medical contexts, and its full potential can be realized through specialized medical-scene methods like our MedSem-DeID.

In Tab.~\ref{tab:result_medical_utility} (3rd column), we summarize the priors employed by different methods.
Landmark priors (DeepPrivacy and CIAGAN) and high-level common priors (face attributes/StyleGAN adopted by Personal/RiDDLE) perform poorly in medical applications, \textit{i.e.}, less than 50\% accuracy and 0.2 segmentation Dice score. The gaze prior (Disguise) is effective for coarse-grained classification (76.01\%) but fails in fine-grained segmentation task (0.2751). Password, using a U-Net to preserve high-frequency signals, excels in low-level segmentation (0.6336) but fails in high-level classification task (54.21\%). This also introduces visual artifacts (Fig.~\ref{fig:qua_benchmark}, 4rd column). In contrast, our method, leveraging the medical semantics within MeMa, achieves superior performance in both classification (86.70\%) and segmentation (0.6775) without handcrafted designs such as landmark.

\noindent \textbf{Real-World Clinical Utility.} We conduct a clinical trial on the Real-ECXHCSU cohort.
As shown in Tab.~\ref{tab:cohens_kappa}, our method largely outperforms the recent DeID methods (Disguise and Password) across all five disease categories, achieving near-perfect consistency in the clinical outcomes, \textit{i.e.}, $k \textgreater 0.81$.
Moreover, our approach is more flexible and effective than a recent hand-crafted DeID approach that is delicately designed for eye diseases, namely, digital mask (DM)~\cite{yang2022digital}.
On complex diseases, such as BCC and Eyelid Nevus (EyelidN), DM does not work ($k$ = 0.0566/0.0988) while our approach achieves satisfactory results ($k$ = 0.8245/0.8346).
Moreover, this real-world evaluation introduces additional disease categories not seen during training, \textit{i.e.}, Entropion and EyelidN, highlighting the robustness and generalizability of our method.
The diagnosis accuracy is provided in the {supplementary material}.

\begin{table}[!thbp]
	\centering
	\renewcommand{\arraystretch}{1.00}
	\setlength{\tabcolsep}{3.5pt} 
	\begin{tabular}{c |ccccc}
		\hline
	\multirow{2}{*}{Method}	& \multicolumn{5}{c}{Cohen's Kappa ($k$) $\uparrow$} \\
		\cline{2-6}
		&BCC&TAO&Ptosis&Entropion&EyelidN \\
		\hline
		\hline
		DM &  {0.0566} & {0.8159}&{0.8276}&{0.1879}&{0.0988}\\
		Disguise& {0.7534} & {0.5824}&{0.7134}&{0.2467}&{0.1387}\\
		Password& {0.4657} & {0.2758}&{0.4289}&{0.1329}&{0.0459}\\
		RiDDLE &  {0.1201} & {0.0751}&{0.0826}&{0.0937}&{0.0811}\\
		Ours& \best{0.8245} & \best{0.8278}&\best{0.8302}&\best{0.8256}&\best{0.8346}\\
		\hline
	\end{tabular}
	\vspace{-2mm}
	\caption{
		Comparison of different DeID methods in terms of the diagnosis outcomes, on the real-world cohort Real-ECXHCSU. $k\textgreater 0.81$ indicates perfect clinical consistency.
	}
	\label{tab:cohens_kappa}
	\vspace{-3mm}
\end{table}

\begin{table}[!thbp]
	\centering
	\renewcommand{\arraystretch}{1.00}
	\setlength{\tabcolsep}{3.8pt} 
	\begin{tabular}{c |c |ccc}
		\hline
		\multirow{2}{*}{Method} &\multirow{2}{*}{Rev}& \multicolumn{3}{c}{ID-Dis $\uparrow$} \\
		\cline{3-5}
		&& FaceNet$_{\scalebox{0.5}{VGGFace2}}$ & FaceNet$_{\scalebox{0.5}{CASIA}}$& Sphere \\
		\hline
		\hline
		DeepPrivacy&\xmark&1.1548&1.1831&1.1818\\
		
		CIAGAN&\xmark&1.2843& 1.2566&1.2881\\
		Disguise&\xmark&1.3976&1.3607&1.3128\\
		\hline
		\hline
		Password&\cmark& 1.3380&1.3139&1.2629\\
		Personal&\cmark&1.2819&1.2944&1.2351\\
		RiDDLE&\cmark&\best{1.4278}&\best{1.3694}&\secondbest{1.3583}\\
		Ours&\cmark& \secondbest{1.4007} & \secondbest{1.3609}&\best{1.3601}\\
		\hline
	\end{tabular}
	\vspace{-2mm}
	\caption{
		Comparison of different methods on the MeMa validation set.
		The higher the ID-Dis, the better de-identified.
		\best{Bold} and \secondbest{italic} indicates the best and the second-best result.
	}
	\vspace{-2mm}
	\label{tab:id-dis}
\end{table}

\begin{table}[!thbp]
	\centering
	\renewcommand{\arraystretch}{1.00}
	\setlength{\tabcolsep}{1.2pt} 
	\begin{tabular}{c|ccccc}
		\hline
		Method&DeepPrivacy&Password&Disguise&RiDDLE&Ours \\
		
		\hline
		Rate(\%) $\downarrow$&5.76&7.76& 2.89& 2.13&\best{1.76} \\
		\hline
	\end{tabular}
	\vspace{-2mm}
	\caption{
		Face matching rate on Real-ECXHCSU.
	}
	\vspace{-6mm}
	\label{tab:id-realworld}
\end{table}

\noindent \textbf{De-Identification Performance.}
As shown in Tab.~\ref{tab:id-dis}, our method achieves the best DeID performance of ID-Dis value 1.3601, when evaluated with the SphereFace face recognition model. 
With the FaceNet$_{\scalebox{0.5}{\text{VGGFace2}}}$ and FaceNet$_{\scalebox{0.5}{\text{CASIA}}}$ models, our approach outperforms all methods except RiDDLE.
RiDDLE maps person images into the very low-dimensional StyleGAN~\cite{karras2019style} latent space and selects a sample with the maximum identity distance from this space. 
While this over-dimension-reduction operation benefits identity protection, it sacrifices much original face information, resulting in poor downstream utilities, as evidenced in Tab.~\ref{tab:result_medical_utility} and Tab.~\ref{tab:cohens_kappa}.

Furthermore, we evaluate our approach on the LFW dataset~\cite{2008Labeled}. With the SphereFace facial recognition network, our approach achieves a face verification accuracy close to random guessing (50\%).

Moreover, 
we simulate a real-world identity authentication system.
We use ID-card photos of Real-ECXHCSU patients as the identity database.
We then match the de-identified clinical photos within the ID photo database.
Note that the ID photo may be a long time away from the clinical photo.
As shown in Tab.~\ref{tab:id-realworld},
our approach achieves the lowest successful face matching rate of 1.76\%, compared to other approaches such as Disguise (2.89\%) and RiDDLE (2.13\%).

\noindent \textbf{Qualitative Results.}
As shown in Fig.~\ref{fig:qua_benchmark}, our approach uniquely preserves both coarse- and fine-grained medical cues, such as drooping eyelids and conjunctival redness. In contrast, DeepPrivacy masks and replaces the original face, Password retains color but distorts shapes. Disguise, Personal, and RiDDLE sacrifice medical cues for privacy. 
Besides, our approach demonstrates good visual quality.

\noindent \textbf{Other Downstream Utilities.}
As shown in Tab.~\ref{tab:other_utilities}, our approach shows competitive or best results on facial landmark detection and gaze detection tasks.
For example, our approach achieves an eye landmark detection error of 2.94, much lower than the second-best approach, Password, which achieves 4.38. This is due to our particularly preserved eye-related medical semantics.
For gaze detection, although Disguise explicitly introduces the gaze detection loss, it still obtains a larger gaze error of 4.85 \textit{v.s.} 3.96, proving the power of our semantics learned on MeMa.

\begin{table}[!thbp]
	\vspace{-0mm}
	\centering
	\renewcommand{\arraystretch}{1.00}
	\setlength{\tabcolsep}{1.4pt} 
	\begin{tabular}{c |c |cccc| cc }
		\hline
		\multirow{2}{*}{Method} &\multirow{2}{*}{Rev}&  \multicolumn{4}{c|}{Landmark Error $\downarrow$} & 
		\multicolumn{2}{c}{Gaze Error $\downarrow$}\\
		\cline{3-8}
		&& All & Eye &Mouth &Nose 
		& Pitch & Yaw 
		\\
		\hline
		\hline
		DeepPrivacy&\xmark&193.91&5.86&155.02&33.02&
		7.72 & 7.06 \\
		
		CIAGAN&\xmark& 313.36&23.87&215.11&74.37&
		13.00&7.78\\

		Disguise&\xmark& 94.09& 5.78&67.29&21.01&
		\secondbest{4.85}&\secondbest{5.71}\\
		\hline
		\hline
		Password&\cmark& \best{65.90}&\secondbest{4.38}&\best{42.52}&\secondbest{18.98}&
		5.34&9.97\\
		
		Personal&\cmark&109.46&5.87&71.20&32.38&
		7.24 & 7.45  \\
		
		RiDDLE&\cmark&136.79&5.73&91.28&39.77&
		7.01&7.93\\
		
		Ours&\cmark&\secondbest{87.62}&\best{2.94}&\secondbest{66.76}&\best{17.92}&
		\best{3.96}&\best{4.97}\\

		\hline
	\end{tabular}
	\vspace{-2mm}
	\caption{Comparison of different DeID methods, in terms of common utilities, on the MeMa validation set.
		Landmark error is calculated as the averaged pixel distance between the original and the de-identified image.
		\best{Bold} and \secondbest{italic} indicates the best and the second-best performance.
	}
	\label{tab:other_utilities}
\end{table}

\noindent \textbf{Reversibility.}
As shown in Tab.~\ref{tab:result_recovery}, compared with other reversible methods, our method performs better in terms of identity recovery, image fidelity, and perceptual quality, achieving ID-Dis, Peak Signal-to-Noise Ratio (PSNR), and LPIPS~\cite{zhang2018unreasonable} values of 0.5642, 27.02dB, and 0.2098, respectively.
Moreover, our approach exhibits the best disease classification accuracy of 89.34\%, while the second-best Personal only obtains 73.08\%.

\begin{figure}[!thbp]
	\centering
	\includegraphics[width=0.97\columnwidth]{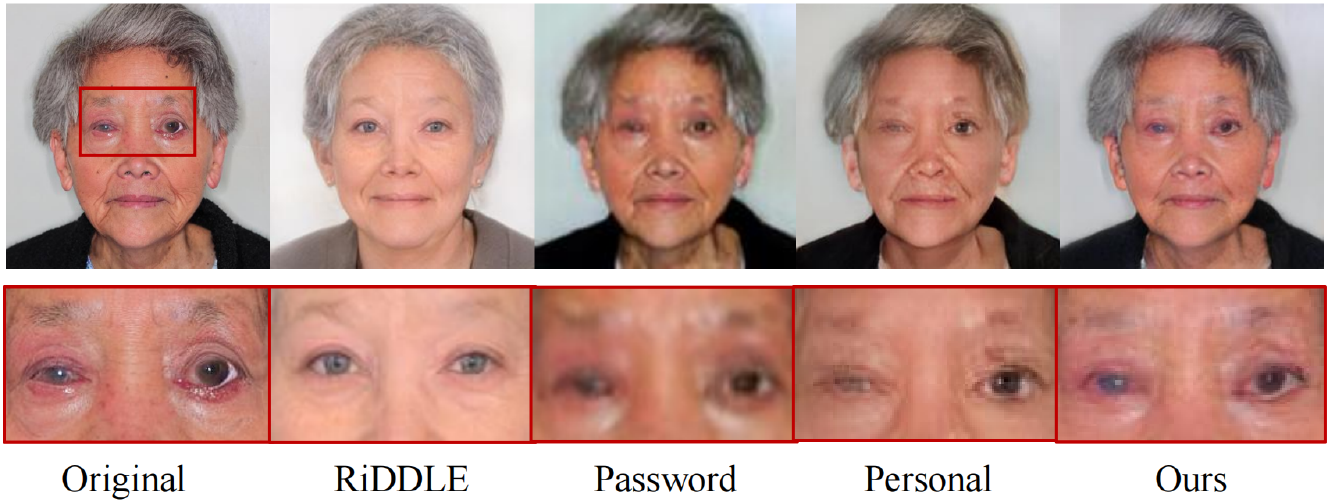}
	\vspace{-2mm}
	\caption{Qualitative results of the recovered image of different reversible DeID approaches.
		The ocular region is zoomed-in for a more clear comparison.
	}
	\vspace{-3mm}
	\label{fig:qua_reverse}
\end{figure}

We provide the qualitative results in Fig.~\ref{fig:qua_reverse}. Only our approach precisely preserves the clinical diagnosis necessary sign, \textit{i.e.}, the discolored left eye iris.
RiDDLE generates high-quality facial textures, while obsoleting medical details.
Password and Personal can not generate sharp details.

\begin{table}[!thbp]
	\centering
	\renewcommand{\arraystretch}{1.00}
	\setlength{\tabcolsep}{4.45pt} 
	\begin{tabular}{c |ccc cH}
		\hline
		{Method}&ID-Dis$\downarrow$& PSNR$\uparrow$ &LPIPS$\downarrow$ & Med-Class$\uparrow$&Med-Seg$\uparrow$\\
		\hline
		\hline
		Password&0.6382&26.29dB&0.2752&67.99\%&-\\
		Personal&0.5723&25.92dB&0.2240&73.08\%&-\\
		RiDDLE&0.8593&14.00dB&0.3732&47.46\%&-\\
		
		Ours&\best{0.5642}&\best{27.02dB}&\best{0.2098}&\best{89.34\%}&\best{}\\
		\hline
	\end{tabular}
	\vspace{-2mm}
	\caption{
		Comparison of the recovered image by various reversible approaches on MaMa validation set.
		Lower ID-Dis indicates the recovered identity is more similar to the original.
		Lower LPIPS indicates better perceptual quality.
	}
	\label{tab:result_recovery}
\end{table}

\subsection{Model Analysis}

\noindent \textbf{Ablation Study for MedSem-DeID Model.}
Recalling that MedSem-DeID enhances the medical knowledge of the DeID pipeline in both the feature extraction procedure ($f_{med}$) and the loss function ($\mathcal{L}_{med}$), we verify the effectiveness of both strategies.
As shown in Tab.~\ref{tab:ablation_framework}, when trained on the common-scene facial dataset FFHQ without using any medical prior, the resulting model (M1) achieves an ID-Dis score of 1.3689 and a disease classification accuracy of 42.13\%.
When the training dataset is replaced with MeMa, the medical accuracy of the resulting model (M2) improves to 46.82\% without compromising the DeID performance.
After further introducing medical priors, no matter the $f_{med}$ or the $\mathcal{L}_{med}$, the resulting models M3 and M4 show an obvious improvement in classification accuracy, \textit{i.e.}, 69.94\% and 71.35\%, while slightly compromising the DeID results.
When combining both strategies, the final model achieves a strong medical performance of 86.70\%.

\begin{table}[!thbp]
	\centering
	\renewcommand{\arraystretch}{1.00}
	\setlength{\tabcolsep}{4.9pt} 
	\begin{tabular}{c|c|c |c |c c}
		\hline
		Model&Dataset&$f_{med}$&$\mathcal{L}_{med}$&{ID-Dis}$\uparrow$ &Med-Class$\uparrow$\\
		\hline
		\hline
		M1&FFHQ&\xmark & \xmark & 1.3609 &42.13\%\\
		M2&MeMa&\xmark & \xmark & 1.3609 &46.82\%\\
		M3&MeMa&\cmark & \xmark & 1.3602 & 69.94\%\\
		M4&MeMa&\xmark & \cmark & 1.3601 & 71.35\%\\
		
		Ours&MeMa&\cmark & \cmark & 1.3601 &\best{86.70\%}\\
		\hline
	\end{tabular}
	\vspace{-2mm}
	\caption{
		Framework ablation Study. Both the MeMa dataset and the utilization of medical priors are useful.
		ID-Dis is calculated with the SphereFace network.
		Med-Class denotes the disease classification accuracy.
	}
	\label{tab:ablation_framework}
\end{table}

\noindent \textbf{Different Medical Semantic Encoders.}
Our method is flexible, not relying on the typical implementation of the medical encoder. To verify this, we trained two other semantic encoders on MeMa. We fine-tuned a pre-trained ViT model~\cite{sharir2021image} using supervised and self-supervised learning strategies, specifically the masked auto-encoder (MAE)~\cite{he2022masked}.
As shown in Tab.~\ref{tab:ablation_encoder}, all three variants achieved decent performance. The supervised ViT performed the poorest due to the sparse disease category label for supervision. 
The diffusion model outperformed ViT(MAE) with 86.38\% \textit{vs.} 85.26\%. This is likely because the LAION-5B~\cite{schuhmann2022laion} dataset used for pre-training the base stable diffusion model is much larger than the pre-training dataset for the base ViT.

\begin{table}[!thbp]
	\centering
	\renewcommand{\arraystretch}{1.00}
	\setlength{\tabcolsep}{4.8pt} 
	\begin{tabular}{c|ccc}
		\hline
		&ViT(Supervised)&ViT(MAE) &Diffusion\\
		\hline
		
		Med-Class$\uparrow$&82.97\%&85.26\%&\best{86.70\%}\\
		\hline
		ID-Dis$\uparrow$&1.3600&1.3601&{1.3601}\\
		\hline
	\end{tabular}
	\vspace{-2mm}
	\caption{
		Impact of various medical semantic encoders.
	}
	\vspace{-2mm}
	\label{tab:ablation_encoder}
\end{table}

\begin{figure}[!thbp]
	\centering
	\begin{tabular}{cc}
		\setlength{\tabcolsep}{0pt} 
		\hspace{-3mm}\includegraphics[width=0.505\columnwidth]{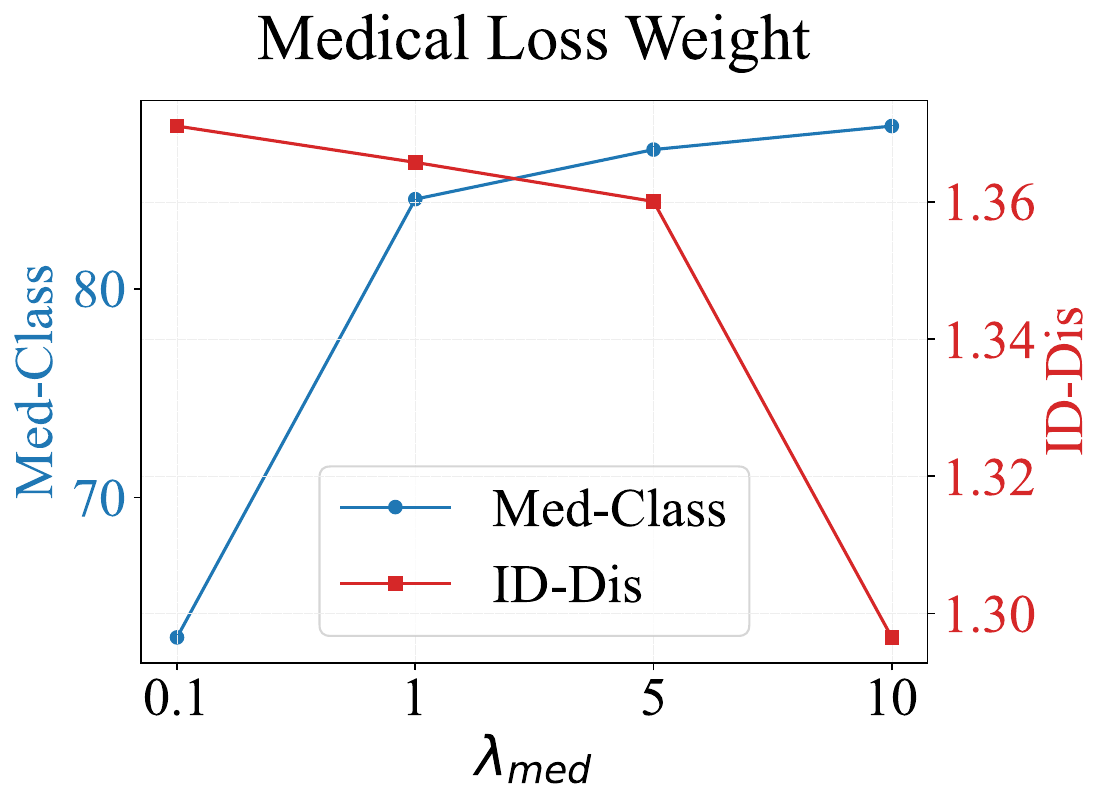}&\hspace{-4mm}
		\includegraphics[width=0.49\columnwidth]{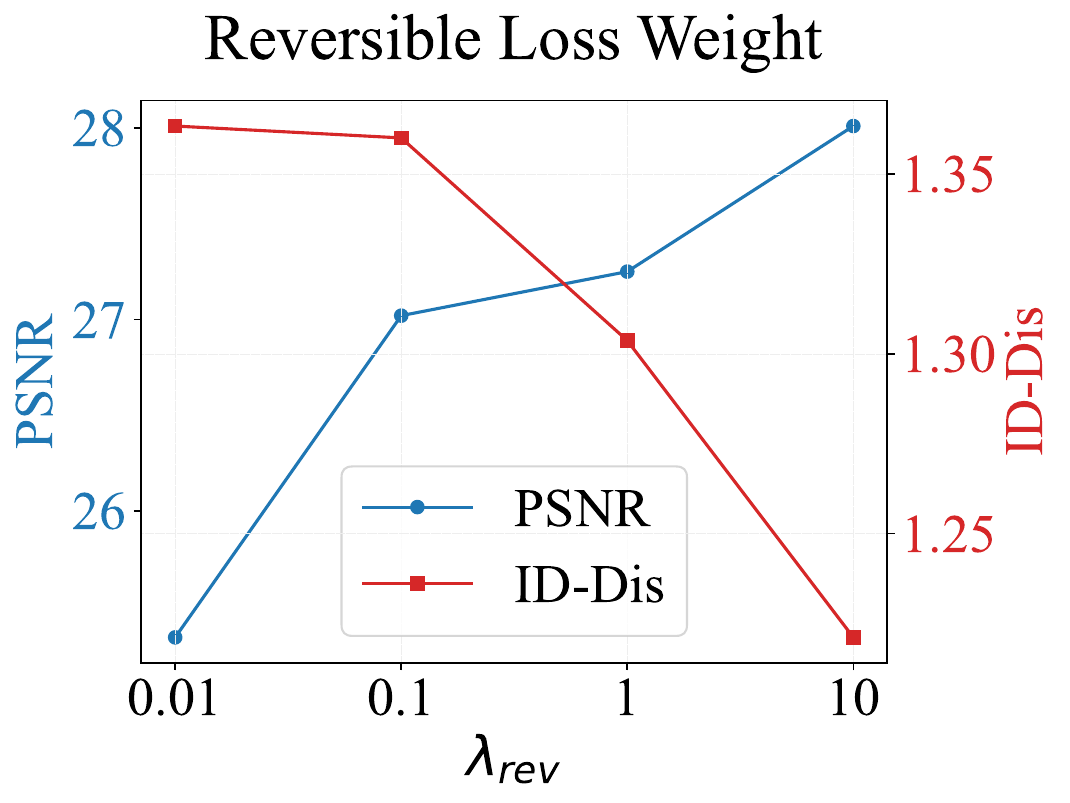}\\
	\end{tabular}
	\vspace{-3mm}
	\caption{Impact of the loss weights for the medical  (\textit{left}) and reversible (\textit{right}) loss terms.} 
	\vspace{-2mm}
	\label{fig:exp_ablation_loss}
\end{figure}

\noindent \textbf{Different Loss Weights.}
As shown in Fig.~\ref{fig:exp_ablation_loss} (\textit{left}), increasing the weight of medical loss (\(\lambda_{med}\)) consistently improves disease classification accuracy, due to the enhanced medical information. However, this also makes the de-identified images more similar to the originals, compromising the DeID performance, \textit{i.e.}, the reduced ID-Dis.
We set \(\lambda_{med}=5\) to achieve the best trade-off between medical accuracy and DeID.
For the weight controlling reversible reconstruction (\(\lambda_{rev}\)), a similar trade-off between the reconstructed image quality and DeID performance is observed, as shown in Fig.~\ref{fig:exp_ablation_loss} (\textit{right}). We set \(\lambda_{rev} = 0.1\) to achieve optimal results.

\vspace{-3mm}
\section{Conclusion and Limitation}
We have released a large-scale patient face dataset, {MeMa}, to facilitate research on medical privacy protection. Expert physicians validated and annotated MeMa. On this dataset, we established a comprehensive benchmark for medical-scene de-identification, also proposing a new baseline approach that outperforms previous approaches.
A limitation is that the current dataset focuses on eye disease-related manifestations. Future work will expand the dataset to include other facial diseases, such as facial paralysis.

\vspace{-3mm}
\section{Acknowledgment}
This work is supported by Strategic Research and Consulting Project of Chinese Academy of Engineering (2024-XBZD-18), National Natural Science Foundation of China (62225112), Shanghai Artificial Intelligence Laboratory, National Natural Science Foundation of China (62101326), National Natural Science Foundation of China (82388101), National Natural Science Foundation of China (72293585), and National Natural Science Foundation of China (72293580). We thank Min Zhou (Doctor of Medicine) and Xuefei Song (Doctor of Medicine) for their invaluable assistance with patient data collection, data annotation, and medical knowledge support.

\bibliography{aaai25}

\end{document}

%% file: exp_medical.tex
\begin{figure*}[t]
	\centering
	\includegraphics[width=0.98\linewidth]{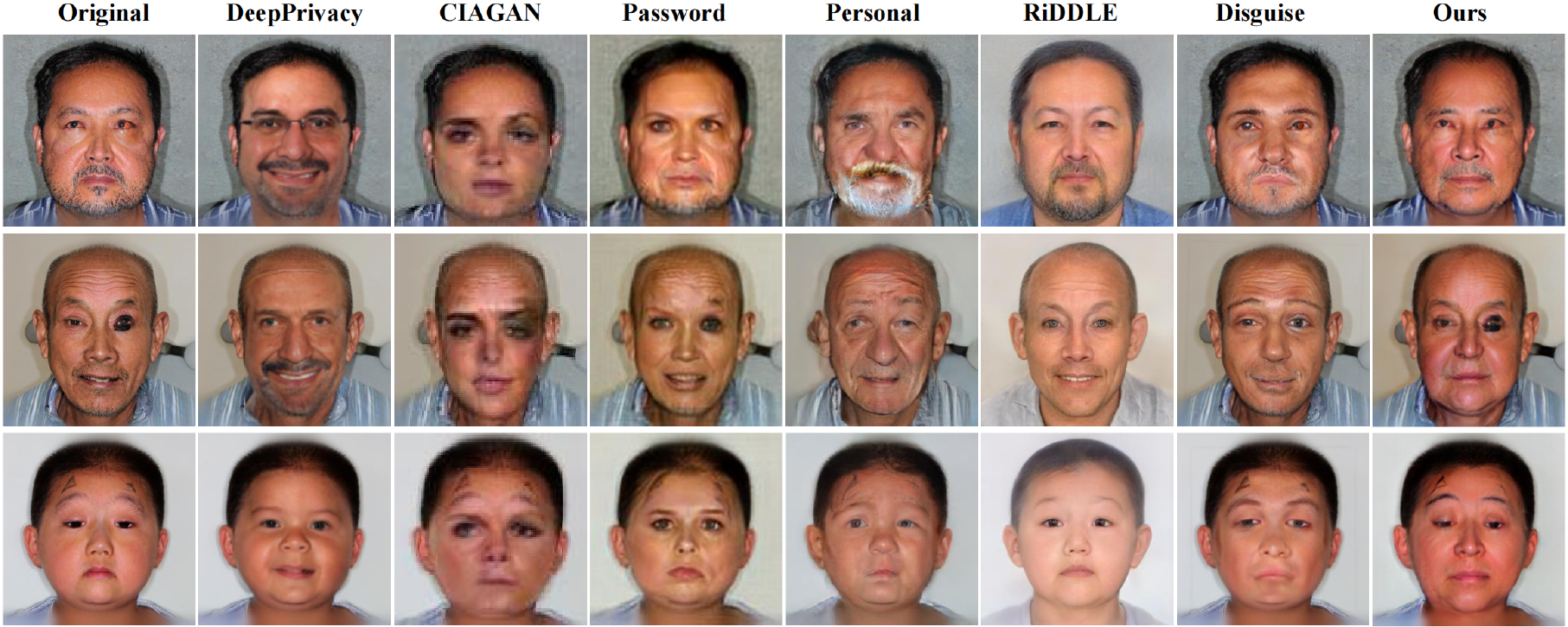}
		\vspace{-2mm}
	\caption{
	Qualitative results of different methods on the MeMA validation set.
	}
		\vspace{-2mm}
	\label{fig:qua_benchmark}
\end{figure*}

\begin{table*}[htbp]
	\centering
	\renewcommand{\arraystretch}{1.00}
	\label{tab:my_label}
	\setlength{\tabcolsep}{3.1pt} 
	\begin{tabular}{c |c |c| ccccccccc|cc}
		\hline
		\multirow{2}{*}{Method} &\multirow{2}{*}{Rev}& \multirow{2}{*}{Utility Prior}& 
		\multicolumn{9}{c|}{Classification (\%) $\uparrow$} &
		\multicolumn{2}{c}{Segmentation $\uparrow$} \\
		\cline{4-14}
		&&&  All & BCC & Conj &Normal & Ptosis& SCC&Strab&TAO&Uveitis&
		Dice& Jaccard\\
		\hline
		\hline
		DeepPrivacy&\xmark&Landmark&40.80 & 91.20 & 4.52 & 80.38 & 69.10 & 0.95 & 3.66 & 59.23 & 17.32 &
		0.0041&0.0021\\
		
		CIAGAN&\xmark&Landmark& 41.36 & 30.96 & 3.85 & 96.93 & 49.36 & 13.52 & 7.33 & 31.81 & 97.11 &
		0.1280&0.0757\\
		Disguise&\xmark&Landmark+Gaze&76.01 & 57.19 & 95.32 & 98.63 & 90.68 & 58.29 & 37.00 & 90.69 & 80.31 &
		0.2751&0.1984
		\\
		\hline
		\hline
		Password&\cmark&Unet&54.21 & 43.99 & 95.99 & 87.37 & 47.53 & 52.38 & 5.93 & 79.26 & 21.22 &
		0.6336&0.5192\\
		Personal&\cmark&Face Attributes&30.17 & 67.17 & 9.69 & 61.95 & 31.99 & 4.19 & 2.09 & 33.39 & 30.90 &
		0.0136&0.0073
		\\
		RiDDLE&\cmark&StyleGAN&30.01 & 29.27 & 0.00 & 74.40 & 58.32 & 0.00 & 0.00 & 4.04 & 74.02 &
		0.0031&0.0017
		\\
		
		Ours&\cmark& Med-Knowledge&\best{86.70} & \best{96.45} & \best{99.83} & \best{98.98} & \best{92.32} & \best{87.05} & \best{63.70} & \best{97.72} & {57.56} &
		\best{0.6775}&\best{0.5453}
		\\
		
		\hline
	\end{tabular}
		\vspace{-2mm}
	\caption{Comparison of various DeID methods on medical tasks.
		Classification and Segmentation tasks are evaluated on validation sets of MeMa and MeMa-Seg, respectively.
		`Rev' denotes if the method is reversible.
	}
	\label{tab:result_medical_utility}
		\vspace{-5mm}
\end{table*}

%% file: AAAI2025.bbl
\begin{thebibliography}{76}
\providecommand{\natexlab}[1]{#1}

\bibitem[{Abdelrahman et~al.(2023)Abdelrahman, Hempel, Khalifa, Al-Hamadi, and
  Dinges}]{abdelrahman2023l2cs}
Abdelrahman, A.~A.; Hempel, T.; Khalifa, A.; Al-Hamadi, A.; and Dinges, L.
  2023.
\newblock L2cs-net: Fine-grained gaze estimation in unconstrained environments.
\newblock In \emph{2023 8th International Conference on Frontiers of Signal
  Processing (ICFSP)}, 98--102. IEEE.

\bibitem[{Banerjee et~al.(1999)Banerjee, Capozzoli, McSweeney, and
  Sinha}]{banerjee1999beyond}
Banerjee, M.; Capozzoli, M.; McSweeney, L.; and Sinha, D. 1999.
\newblock Beyond kappa: A review of interrater agreement measures.
\newblock \emph{Canadian journal of statistics}, 27(1): 3--23.

\bibitem[{Cai et~al.(2024)Cai, Gao, Planche, Zheng, Chen, Asif, and
  Wu}]{cai2024disguise}
Cai, Z.; Gao, Z.; Planche, B.; Zheng, M.; Chen, T.; Asif, M.~S.; and Wu, Z.
  2024.
\newblock Disguise without disruption: Utility-preserving face
  de-identification.
\newblock In \emph{Proceedings of the AAAI Conference on Artificial
  Intelligence}, volume~38, 918--926.

\bibitem[{Cao et~al.(2021)Cao, Liu, Wen, Xie, and Song}]{cao2021personalized}
Cao, J.; Liu, B.; Wen, Y.; Xie, R.; and Song, L. 2021.
\newblock Personalized and invertible face de-identification by disentangled
  identity information manipulation.
\newblock In \emph{Proceedings of the IEEE/CVF international conference on
  computer vision}, 3334--3342.

\bibitem[{Cao et~al.(2018)Cao, Shen, Xie, Parkhi, and
  Zisserman}]{cao2018vggface2}
Cao, Q.; Shen, L.; Xie, W.; Parkhi, O.~M.; and Zisserman, A. 2018.
\newblock Vggface2: A dataset for recognising faces across pose and age.
\newblock In \emph{2018 13th IEEE international conference on automatic face \&
  gesture recognition (FG 2018)}, 67--74. IEEE.

\bibitem[{Chen et~al.(2024{\natexlab{a}})Chen, Qu, Tian, Jiang, Qin, Gao,
  Zhang, Ma, Jin, and Zhai}]{chen2024cross}
Chen, H.; Qu, Z.; Tian, Y.; Jiang, N.; Qin, Y.; Gao, J.; Zhang, R.; Ma, Y.;
  Jin, Z.; and Zhai, G. 2024{\natexlab{a}}.
\newblock A cross-temporal multimodal fusion system based on deep learning for
  orthodontic monitoring.
\newblock \emph{Computers in Biology and Medicine}, 180: 109025.

\bibitem[{Chen et~al.(2024{\natexlab{b}})Chen, Sun, Tian, Jia, Zhang, Wang,
  Huang, Min, Zhai, and Zhang}]{chen2024gaia}
Chen, Z.; Sun, W.; Tian, Y.; Jia, J.; Zhang, Z.; Wang, J.; Huang, R.; Min, X.;
  Zhai, G.; and Zhang, W. 2024{\natexlab{b}}.
\newblock GAIA: Rethinking Action Quality Assessment for AI-Generated Videos.
\newblock \emph{arXiv preprint arXiv:2406.06087}.

\bibitem[{Deng et~al.(2019)Deng, Guo, Xue, and Zafeiriou}]{deng2019arcface}
Deng, J.; Guo, J.; Xue, N.; and Zafeiriou, S. 2019.
\newblock Arcface: Additive angular margin loss for deep face recognition.
\newblock In \emph{Proceedings of the IEEE/CVF conference on computer vision
  and pattern recognition}, 4690--4699.

\bibitem[{Duan et~al.(2022)Duan, Shen, Min, Tian, Jung, Yang, and
  Zhai}]{duan2022develop}
Duan, H.; Shen, W.; Min, X.; Tian, Y.; Jung, J.-H.; Yang, X.; and Zhai, G.
  2022.
\newblock Develop then rival: A human vision-inspired framework for
  superimposed image decomposition.
\newblock \emph{IEEE Transactions on Multimedia}, 25: 4267--4281.

\bibitem[{Fletcher, Islam et~al.(2018)}]{fletcher2018comparing}
Fletcher, S.; Islam, M.~Z.; et~al. 2018.
\newblock Comparing sets of patterns with the Jaccard index.
\newblock \emph{Australasian Journal of Information Systems}, 22.

\bibitem[{Gao et~al.(2024)Gao, Jiang, Wu, Ma, Li, and Liu}]{gao2024imofc}
Gao, C.; Jiang, Y.; Wu, S.; Ma, Y.; Li, L.; and Liu, D. 2024.
\newblock IMOFC: Identity-Level Metric Optimized Feature Compression for
  Identification Tasks.
\newblock \emph{IEEE Transactions on Circuits and Systems for Video
  Technology}.

\bibitem[{Gao et~al.(2022)Gao, Li, Liu, Chen, Li, and Wu}]{gao2022two}
Gao, C.; Li, L.; Liu, D.; Chen, Z.; Li, W.; and Wu, F. 2022.
\newblock Two-step fast mode decision for intra coding of screen content.
\newblock \emph{IEEE Transactions on Circuits and Systems for Video
  Technology}, 32(8): 5608--5622.

\bibitem[{Gao et~al.(2021)Gao, Liu, Li, and Wu}]{gao2021towards}
Gao, C.; Liu, D.; Li, L.; and Wu, F. 2021.
\newblock Towards task-generic image compression: A study of semantics-oriented
  metrics.
\newblock \emph{IEEE Transactions on Multimedia}, 25: 721--735.

\bibitem[{Gross et~al.(2006)Gross, Sweeney, De~la Torre, and
  Baker}]{gross2006model}
Gross, R.; Sweeney, L.; De~la Torre, F.; and Baker, S. 2006.
\newblock Model-based face de-identification.
\newblock In \emph{2006 Conference on computer vision and pattern recognition
  workshop (CVPRW'06)}, 161--161. IEEE.

\bibitem[{Gu et~al.(2020)Gu, Luo, Ryoo, and Lee}]{gu2020password}
Gu, X.; Luo, W.; Ryoo, M.~S.; and Lee, Y.~J. 2020.
\newblock Password-conditioned anonymization and deanonymization with face
  identity transformers.
\newblock In \emph{European conference on computer vision}, 727--743. Springer.

\bibitem[{Guideline(2001)}]{guideline2001guideline}
Guideline, I. H.~T. 2001.
\newblock Guideline for good clinical practice.
\newblock \emph{J Postgrad Med}, 47(3): 199--203.

\bibitem[{He et~al.(2022)He, Chen, Xie, Li, Doll{\'a}r, and
  Girshick}]{he2022masked}
He, K.; Chen, X.; Xie, S.; Li, Y.; Doll{\'a}r, P.; and Girshick, R. 2022.
\newblock Masked autoencoders are scalable vision learners.
\newblock In \emph{Proceedings of the IEEE/CVF conference on computer vision
  and pattern recognition}, 16000--16009.

\bibitem[{He et~al.(2020)He, Fan, Wu, Xie, and Girshick}]{he2020momentum}
He, K.; Fan, H.; Wu, Y.; Xie, S.; and Girshick, R. 2020.
\newblock Momentum contrast for unsupervised visual representation learning.
\newblock In \emph{Proceedings of the IEEE/CVF conference on computer vision
  and pattern recognition}, 9729--9738.

\bibitem[{He et~al.(2016)He, Zhang, Ren, and Sun}]{he2016deep}
He, K.; Zhang, X.; Ren, S.; and Sun, J. 2016.
\newblock Deep residual learning for image recognition.
\newblock In \emph{Proceedings of the IEEE conference on computer vision and
  pattern recognition}, 770--778.

\bibitem[{Hedlin et~al.(2024)Hedlin, Sharma, Mahajan, Isack, Kar, Tagliasacchi,
  and Yi}]{hedlin2024unsupervised}
Hedlin, E.; Sharma, G.; Mahajan, S.; Isack, H.; Kar, A.; Tagliasacchi, A.; and
  Yi, K.~M. 2024.
\newblock Unsupervised semantic correspondence using stable diffusion.
\newblock \emph{Advances in Neural Information Processing Systems}, 36.

\bibitem[{Hu et~al.(2021)Hu, Shen, Wallis, Allen-Zhu, Li, Wang, Wang, and
  Chen}]{hu2021lora}
Hu, E.~J.; Shen, Y.; Wallis, P.; Allen-Zhu, Z.; Li, Y.; Wang, S.; Wang, L.; and
  Chen, W. 2021.
\newblock Lora: Low-rank adaptation of large language models.
\newblock \emph{arXiv preprint arXiv:2106.09685}.

\bibitem[{Hu et~al.(2020)Hu, Li, Zhang, Yi, Wang, and Manocha}]{hu2020dgaze}
Hu, Z.; Li, S.; Zhang, C.; Yi, K.; Wang, G.; and Manocha, D. 2020.
\newblock Dgaze: Cnn-based gaze prediction in dynamic scenes.
\newblock \emph{IEEE transactions on visualization and computer graphics},
  26(5): 1902--1911.

\bibitem[{Huang et~al.(2008{\natexlab{a}})Huang, Mattar, Berg, and
  Learned-Miller}]{huang2008labeled}
Huang, G.~B.; Mattar, M.; Berg, T.; and Learned-Miller, E. 2008{\natexlab{a}}.
\newblock Labeled faces in the wild: A database forstudying face recognition in
  unconstrained environments.
\newblock In \emph{Workshop on faces in'Real-Life'Images: detection, alignment,
  and recognition}.

\bibitem[{Huang et~al.(2008{\natexlab{b}})Huang, Mattar, Berg, and
  Learned-Miller}]{2008Labeled}
Huang, G.~B.; Mattar, M.; Berg, T.; and Learned-Miller, E. 2008{\natexlab{b}}.
\newblock Labeled Faces in the Wild: A Database forStudying Face Recognition in
  Unconstrained Environments.
\newblock \emph{Month}.

\bibitem[{Hukkel{\aa}s, Mester, and Lindseth(2019)}]{hukkelaas2019deepprivacy}
Hukkel{\aa}s, H.; Mester, R.; and Lindseth, F. 2019.
\newblock Deepprivacy: A generative adversarial network for face anonymization.
\newblock In \emph{International symposium on visual computing}, 565--578.
  Springer.

\bibitem[{Isensee et~al.(2021)Isensee, Jaeger, Kohl, Petersen, and
  Maier-Hein}]{isensee2021nnu}
Isensee, F.; Jaeger, P.~F.; Kohl, S.~A.; Petersen, J.; and Maier-Hein, K.~H.
  2021.
\newblock nnU-Net: a self-configuring method for deep learning-based biomedical
  image segmentation.
\newblock \emph{Nature methods}, 18(2): 203--211.

\bibitem[{Jourabloo, Yin, and Liu(2015)}]{jourabloo2015attribute}
Jourabloo, A.; Yin, X.; and Liu, X. 2015.
\newblock Attribute preserved face de-identification.
\newblock In \emph{2015 International conference on biometrics (ICB)},
  278--285. IEEE.

\bibitem[{Kamnitsas et~al.(2017)Kamnitsas, Ledig, Newcombe, Simpson, Kane,
  Menon, Rueckert, and Glocker}]{kamnitsas2017efficient}
Kamnitsas, K.; Ledig, C.; Newcombe, V.~F.; Simpson, J.~P.; Kane, A.~D.; Menon,
  D.~K.; Rueckert, D.; and Glocker, B. 2017.
\newblock Efficient multi-scale 3D CNN with fully connected CRF for accurate
  brain lesion segmentation.
\newblock \emph{Medical image analysis}, 36: 61--78.

\bibitem[{Karkkainen and Joo(2021)}]{karkkainen2021fairface}
Karkkainen, K.; and Joo, J. 2021.
\newblock Fairface: Face attribute dataset for balanced race, gender, and age
  for bias measurement and mitigation.
\newblock In \emph{Proceedings of the IEEE/CVF winter conference on
  applications of computer vision}, 1548--1558.

\bibitem[{Karras et~al.(2017)Karras, Aila, Laine, and
  Lehtinen}]{karras2017progressive}
Karras, T.; Aila, T.; Laine, S.; and Lehtinen, J. 2017.
\newblock Progressive growing of gans for improved quality, stability, and
  variation.
\newblock \emph{arXiv preprint arXiv:1710.10196}.

\bibitem[{Karras, Laine, and Aila(2019)}]{karras2019style}
Karras, T.; Laine, S.; and Aila, T. 2019.
\newblock A style-based generator architecture for generative adversarial
  networks.
\newblock In \emph{Proceedings of the IEEE/CVF conference on computer vision
  and pattern recognition}, 4401--4410.

\bibitem[{King(2009)}]{king2009dlib}
King, D.~E. 2009.
\newblock Dlib-ml: A machine learning toolkit.
\newblock \emph{The Journal of Machine Learning Research}, 10: 1755--1758.

\bibitem[{Kingma(2014)}]{kingma2014adam}
Kingma, D. 2014.
\newblock Adam: a method for stochastic optimization.
\newblock \emph{arXiv preprint arXiv:1412.6980}.

\bibitem[{Kirillov et~al.(2023)Kirillov, Mintun, Ravi, Mao, Rolland, Gustafson,
  Xiao, Whitehead, Berg, Lo et~al.}]{kirillov2023segment}
Kirillov, A.; Mintun, E.; Ravi, N.; Mao, H.; Rolland, C.; Gustafson, L.; Xiao,
  T.; Whitehead, S.; Berg, A.~C.; Lo, W.-Y.; et~al. 2023.
\newblock Segment anything.
\newblock In \emph{Proceedings of the IEEE/CVF International Conference on
  Computer Vision}, 4015--4026.

\bibitem[{Kong and Fu(2022)}]{kong2022human}
Kong, Y.; and Fu, Y. 2022.
\newblock Human action recognition and prediction: A survey.
\newblock \emph{International Journal of Computer Vision}, 130(5): 1366--1401.

\bibitem[{Li et~al.(2024{\natexlab{a}})Li, Zhang, Zhang, Wu, Tian, Sun, Lu,
  Liu, Min, Lin et~al.}]{li2024r}
Li, C.; Zhang, J.; Zhang, Z.; Wu, H.; Tian, Y.; Sun, W.; Lu, G.; Liu, X.; Min,
  X.; Lin, W.; et~al. 2024{\natexlab{a}}.
\newblock R-Bench: Are your Large Multimodal Model Robust to Real-world
  Corruptions?
\newblock \emph{arXiv preprint arXiv:2410.05474}.

\bibitem[{Li et~al.(2023)Li, Wang, Zhao, Dong, and Tan}]{li2023riddle}
Li, D.; Wang, W.; Zhao, K.; Dong, J.; and Tan, T. 2023.
\newblock RiDDLE: Reversible and Diversified De-Identification With Latent
  Encryptor.
\newblock In \emph{Proceedings of the IEEE/CVF Conference on Computer Vision
  and Pattern Recognition}, 8093--8102.

\bibitem[{Li et~al.(2024{\natexlab{b}})Li, Lu, Han, and
  Prisacariu}]{li2024sd4match}
Li, X.; Lu, J.; Han, K.; and Prisacariu, V.~A. 2024{\natexlab{b}}.
\newblock Sd4match: Learning to prompt stable diffusion model for semantic
  matching.
\newblock In \emph{Proceedings of the IEEE/CVF Conference on Computer Vision
  and Pattern Recognition}, 27558--27568.

\bibitem[{Liu et~al.(2017)Liu, Wen, Yu, Li, Raj, and Song}]{liu2017sphereface}
Liu, W.; Wen, Y.; Yu, Z.; Li, M.; Raj, B.; and Song, L. 2017.
\newblock Sphereface: Deep hypersphere embedding for face recognition.
\newblock In \emph{Proceedings of the IEEE conference on computer vision and
  pattern recognition}, 212--220.

\bibitem[{Maximov, Elezi, and Leal-Taix{\'e}(2020)}]{maximov2020ciagan}
Maximov, M.; Elezi, I.; and Leal-Taix{\'e}, L. 2020.
\newblock Ciagan: Conditional identity anonymization generative adversarial
  networks.
\newblock In \emph{Proceedings of the IEEE/CVF conference on computer vision
  and pattern recognition}, 5447--5456.

\bibitem[{Merck \&~Co.(2024)}]{msdmanuals2024}
Merck \&~Co., R. N.~U., Inc. 2024.
\newblock MSD MANUALS: The Trusted Provider of Medical Information since 1899.
\newblock \url{https://www.msdmanuals.com/}.

\bibitem[{Min et~al.(2024)Min, Duan, Sun, Zhu, and Zhai}]{min2024perceptual}
Min, X.; Duan, H.; Sun, W.; Zhu, Y.; and Zhai, G. 2024.
\newblock Perceptual video quality assessment: A survey.
\newblock \emph{Science China Information Sciences}, 67(11): 211301.

\bibitem[{Mohsenin and Huang(2012)}]{mohsenin2012ocular}
Mohsenin, A.; and Huang, J.~J. 2012.
\newblock Ocular manifestations of systemic inflammatory diseases.
\newblock \emph{Connecticut medicine}, 76(9).

\bibitem[{Oquab et~al.(2023)Oquab, Darcet, Moutakanni, Vo, Szafraniec,
  Khalidov, Fernandez, Haziza, Massa, El-Nouby et~al.}]{oquab2023dinov2}
Oquab, M.; Darcet, T.; Moutakanni, T.; Vo, H.; Szafraniec, M.; Khalidov, V.;
  Fernandez, P.; Haziza, D.; Massa, F.; El-Nouby, A.; et~al. 2023.
\newblock Dinov2: Learning robust visual features without supervision.
\newblock \emph{arXiv preprint arXiv:2304.07193}.

\bibitem[{Price and Cohen(2019)}]{price2019privacy}
Price, W.~N.; and Cohen, I.~G. 2019.
\newblock Privacy in the age of medical big data.
\newblock \emph{Nature medicine}, 25(1): 37--43.

\bibitem[{Ren, Lee, and Ryoo(2018)}]{ren2018learning}
Ren, Z.; Lee, Y.~J.; and Ryoo, M.~S. 2018.
\newblock Learning to anonymize faces for privacy preserving action detection.
\newblock In \emph{Proceedings of the european conference on computer vision
  (ECCV)}, 620--636.

\bibitem[{Rombach et~al.(2022)Rombach, Blattmann, Lorenz, Esser, and
  Ommer}]{rombach2022high}
Rombach, R.; Blattmann, A.; Lorenz, D.; Esser, P.; and Ommer, B. 2022.
\newblock High-resolution image synthesis with latent diffusion models.
\newblock In \emph{Proceedings of the IEEE/CVF conference on computer vision
  and pattern recognition}, 10684--10695.

\bibitem[{R{\"u}schendorf(1985)}]{ruschendorf1985wasserstein}
R{\"u}schendorf, L. 1985.
\newblock The Wasserstein distance and approximation theorems.
\newblock \emph{Probability Theory and Related Fields}, 70(1): 117--129.

\bibitem[{Schroff, Kalenichenko, and Philbin(2015)}]{schroff2015facenet}
Schroff, F.; Kalenichenko, D.; and Philbin, J. 2015.
\newblock Facenet: A unified embedding for face recognition and clustering.
\newblock In \emph{Proceedings of the IEEE conference on computer vision and
  pattern recognition}, 815--823.

\bibitem[{Schuhmann et~al.(2022)Schuhmann, Beaumont, Vencu, Gordon, Wightman,
  Cherti, Coombes, Katta, Mullis, Wortsman et~al.}]{schuhmann2022laion}
Schuhmann, C.; Beaumont, R.; Vencu, R.; Gordon, C.; Wightman, R.; Cherti, M.;
  Coombes, T.; Katta, A.; Mullis, C.; Wortsman, M.; et~al. 2022.
\newblock Laion-5b: An open large-scale dataset for training next generation
  image-text models.
\newblock \emph{Advances in Neural Information Processing Systems}, 35:
  25278--25294.

\bibitem[{Serengil(2020)}]{serengil2020deepface}
Serengil, S. 2020.
\newblock DeepFace: Face Recognition with Deep Neural Networks.
\newblock \url{https://github.com/serengil/deepface}.

\bibitem[{Serengil and Ozpinar(2021)}]{serengil2021lightface}
Serengil, S.~I.; and Ozpinar, A. 2021.
\newblock HyperExtended LightFace: A Facial Attribute Analysis Framework.
\newblock In \emph{2021 International Conference on Engineering and Emerging
  Technologies (ICEET)}, 1--4. IEEE.

\bibitem[{Sharir, Noy, and Zelnik-Manor(2021)}]{sharir2021image}
Sharir, G.; Noy, A.; and Zelnik-Manor, L. 2021.
\newblock An image is worth 16x16 words, what is a video worth?
\newblock \emph{arXiv preprint arXiv:2103.13915}.

\bibitem[{Tan et~al.(2024)Tan, Zhu, Cheng, Hu, Zhang, Pei, Yu, Li, Li, and
  Wang}]{tan2024low}
Tan, X.; Zhu, Y.; Cheng, Z.; Hu, M.; Zhang, X.; Pei, G.; Yu, C.; Li, Q.; Li,
  W.; and Wang, J. 2024.
\newblock Low-cost and portable physiological signal monitor using PhysRate
  model.
\newblock \emph{Displays}, 81: 102605.

\bibitem[{Tang et~al.(2023)Tang, Jia, Wang, Phoo, and
  Hariharan}]{tang2023emergent}
Tang, L.; Jia, M.; Wang, Q.; Phoo, C.~P.; and Hariharan, B. 2023.
\newblock Emergent correspondence from image diffusion.
\newblock \emph{Advances in Neural Information Processing Systems}, 36:
  1363--1389.

\bibitem[{Tian et~al.(2024{\natexlab{a}})Tian, Aggarwal, Colaco, Kira, and
  Gonzalez-Franco}]{tian2024diffuse}
Tian, J.; Aggarwal, L.; Colaco, A.; Kira, Z.; and Gonzalez-Franco, M.
  2024{\natexlab{a}}.
\newblock Diffuse Attend and Segment: Unsupervised Zero-Shot Segmentation using
  Stable Diffusion.
\newblock In \emph{Proceedings of the IEEE/CVF Conference on Computer Vision
  and Pattern Recognition}, 3554--3563.

\bibitem[{Tian et~al.(2020)Tian, Che, Bao, Zhai, and Gao}]{tian2020self}
Tian, Y.; Che, Z.; Bao, W.; Zhai, G.; and Gao, Z. 2020.
\newblock Self-supervised motion representation via scattering local motion
  cues.
\newblock In \emph{Computer Vision--ECCV 2020: 16th European Conference,
  Glasgow, UK, August 23--28, 2020, Proceedings, Part XIV 16}, 71--89.
  Springer.

\bibitem[{Tian et~al.(2021)Tian, Lu, Min, Che, Zhai, Guo, and
  Gao}]{tian2021self}
Tian, Y.; Lu, G.; Min, X.; Che, Z.; Zhai, G.; Guo, G.; and Gao, Z. 2021.
\newblock Self-conditioned probabilistic learning of video rescaling.
\newblock In \emph{Proceedings of the IEEE/CVF international conference on
  computer vision}, 4490--4499.

\bibitem[{Tian et~al.(2024{\natexlab{b}})Tian, Lu, Yan, Zhai, Chen, and
  Gao}]{tian2024coding}
Tian, Y.; Lu, G.; Yan, Y.; Zhai, G.; Chen, L.; and Gao, Z. 2024{\natexlab{b}}.
\newblock A coding framework and benchmark towards low-bitrate video
  understanding.
\newblock \emph{IEEE Transactions on Pattern Analysis and Machine
  Intelligence}.

\bibitem[{Tian, Lu, and Zhai(2024)}]{tian2024smc++}
Tian, Y.; Lu, G.; and Zhai, G. 2024.
\newblock SMC++: Masked Learning of Unsupervised Video Semantic Compression.
\newblock \emph{arXiv preprint arXiv:2406.04765}.

\bibitem[{Tian, Lu, and Zhai(2025)}]{tian2025free}
Tian, Y.; Lu, G.; and Zhai, G. 2025.
\newblock Free-VSC: Free Semantics from Visual Foundation Models for
  Unsupervised Video Semantic Compression.
\newblock In \emph{European Conference on Computer Vision}, 163--183. Springer.

\bibitem[{Tian et~al.(2023{\natexlab{a}})Tian, Lu, Zhai, and Gao}]{tian2023non}
Tian, Y.; Lu, G.; Zhai, G.; and Gao, Z. 2023{\natexlab{a}}.
\newblock Non-semantics suppressed mask learning for unsupervised video
  semantic compression.
\newblock In \emph{Proceedings of the IEEE/CVF International Conference on
  Computer Vision}, 13610--13622.

\bibitem[{Tian et~al.(2019)Tian, Min, Zhai, and Gao}]{tian2019video}
Tian, Y.; Min, X.; Zhai, G.; and Gao, Z. 2019.
\newblock Video-based early asd detection via temporal pyramid networks.
\newblock In \emph{2019 IEEE International Conference on Multimedia and Expo
  (ICME)}, 272--277. IEEE.

\bibitem[{Tian et~al.(2023{\natexlab{b}})Tian, Yan, Zhai, Chen, and
  Gao}]{tian2023clsa}
Tian, Y.; Yan, Y.; Zhai, G.; Chen, L.; and Gao, Z. 2023{\natexlab{b}}.
\newblock Clsa: a contrastive learning framework with selective aggregation for
  video rescaling.
\newblock \emph{IEEE Transactions on Image Processing}, 32: 1300--1314.

\bibitem[{Tian et~al.(2022)Tian, Yan, Zhai, Guo, and Gao}]{tian2022ean}
Tian, Y.; Yan, Y.; Zhai, G.; Guo, G.; and Gao, Z. 2022.
\newblock Ean: event adaptive network for enhanced action recognition.
\newblock \emph{International Journal of Computer Vision}, 130(10): 2453--2471.

\bibitem[{Vaswani et~al.(2017)Vaswani, Shazeer, Parmar, Uszkoreit, Jones,
  Gomez, Kaiser, and Polosukhin}]{vaswani2017attention}
Vaswani, A.; Shazeer, N.; Parmar, N.; Uszkoreit, J.; Jones, L.; Gomez, A.~N.;
  Kaiser, {\L}.; and Polosukhin, I. 2017.
\newblock Attention is all you need.
\newblock \emph{Advances in neural information processing systems}, 30.

\bibitem[{Wang et~al.(2018)Wang, Wang, Zhou, Ji, Gong, Zhou, Li, and
  Liu}]{wang2018cosface}
Wang, H.; Wang, Y.; Zhou, Z.; Ji, X.; Gong, D.; Zhou, J.; Li, Z.; and Liu, W.
  2018.
\newblock Cosface: Large margin cosine loss for deep face recognition.
\newblock In \emph{Proceedings of the IEEE conference on computer vision and
  pattern recognition}, 5265--5274.

\bibitem[{Wen et~al.(2023)Wen, Liu, Cao, Xie, and Song}]{wen2023divide}
Wen, Y.; Liu, B.; Cao, J.; Xie, R.; and Song, L. 2023.
\newblock Divide and conquer: a two-step method for high quality face
  de-identification with model explainability.
\newblock In \emph{Proceedings of the IEEE/CVF International Conference on
  Computer Vision}, 5148--5157.

\bibitem[{Yan et~al.(2023)Yan, Li, Zhao, Tian, and Zhao}]{yan2023dhbe}
Yan, Z.; Li, S.; Zhao, R.; Tian, Y.; and Zhao, Y. 2023.
\newblock DHBE: data-free holistic backdoor erasing in deep neural networks via
  restricted adversarial distillation.
\newblock In \emph{Proceedings of the 2023 ACM Asia Conference on Computer and
  Communications Security}, 731--745.

\bibitem[{Yang et~al.(2022)Yang, Lyu, Wang, Wen, Zhao, Chen, Bi, Meng, Mao,
  Xiao et~al.}]{yang2022digital}
Yang, Y.; Lyu, J.; Wang, R.; Wen, Q.; Zhao, L.; Chen, W.; Bi, S.; Meng, J.;
  Mao, K.; Xiao, Y.; et~al. 2022.
\newblock A digital mask to safeguard patient privacy.
\newblock \emph{Nature medicine}, 28(9): 1883--1892.

\bibitem[{Ye et~al.(2023)Ye, Zhang, Liu, Han, and Yang}]{ye2023ip}
Ye, H.; Zhang, J.; Liu, S.; Han, X.; and Yang, W. 2023.
\newblock Ip-adapter: Text compatible image prompt adapter for text-to-image
  diffusion models.
\newblock \emph{arXiv preprint arXiv:2308.06721}.

\bibitem[{Yi et~al.(2014)Yi, Lei, Liao, and Li}]{yi2014learning}
Yi, D.; Lei, Z.; Liao, S.; and Li, S.~Z. 2014.
\newblock Learning face representation from scratch.
\newblock \emph{arXiv preprint arXiv:1411.7923}.

\bibitem[{Yi et~al.(2021)Yi, Chen, Sun, Min, Tian, and Zhai}]{yi2021attention}
Yi, F.; Chen, M.; Sun, W.; Min, X.; Tian, Y.; and Zhai, G. 2021.
\newblock Attention based network for no-reference UGC video quality
  assessment.
\newblock In \emph{2021 IEEE international conference on image processing
  (ICIP)}, 1414--1418. IEEE.

\bibitem[{Yi, Jiang, and Zhou(2024)}]{yi2024no}
Yi, X.; Jiang, Q.; and Zhou, W. 2024.
\newblock No-reference quality assessment of underwater image enhancement.
\newblock \emph{Displays}, 81: 102586.

\bibitem[{Zhang et~al.(2024)Zhang, Huang, Jin, and Lu}]{zhang2024vision}
Zhang, J.; Huang, J.; Jin, S.; and Lu, S. 2024.
\newblock Vision-language models for vision tasks: A survey.
\newblock \emph{IEEE Transactions on Pattern Analysis and Machine
  Intelligence}.

\bibitem[{Zhang et~al.(2018)Zhang, Isola, Efros, Shechtman, and
  Wang}]{zhang2018unreasonable}
Zhang, R.; Isola, P.; Efros, A.~A.; Shechtman, E.; and Wang, O. 2018.
\newblock The unreasonable effectiveness of deep features as a perceptual
  metric.
\newblock In \emph{Proceedings of the IEEE conference on computer vision and
  pattern recognition}, 586--595.

\end{thebibliography}
